\documentclass{article}

\usepackage[english]{babel}

\usepackage[letterpaper,top=2cm,bottom=2cm,left=3cm,right=3cm,marginparwidth=1.75cm]{geometry}

\usepackage{natbib}
\usepackage{authblk}

\usepackage{float}
\usepackage{amsmath}
\usepackage{graphicx}
\usepackage{xcolor}
\usepackage[colorlinks=true, allcolors=blue]{hyperref}
\newcommand\Eatdot[1]{}
\title{Fundamentals of Generative Large Language Models\\ and Perspectives in Cyber-Defense}

\author[1, 2, +]{Andrei Kucharavy}
\author[3]{Zachary Schillaci}
\author[1,4]{Loïc Maréchal}
\author[1,5]{Maxime Würsch}
\author[1]{Ljiljana Dolamic}
\author[3]{Remi Sabonnadiere}
\author[1,2]{Dimitri Percia David}
\author[1]{Alain Mermoud}
\author[1]{Vincent Lenders}

\affil[+]{Corresponding Author; andrei.kucharavy@hevs.ch}

\affil[1]{Cyber-Defence Campus, armasuisse S+T}
\affil[2]{Institute of Entrepreneurship \& Management, HES-SO Valais-Wallis}
\affil[3]{Effixis SA}
\affil[4]{Department of Information Systems, HEC Lausanne, University of Lausanne}
\affil[5]{Section of Computer Science, EPFL}

\date{}

\begin{document}
\maketitle
\thispagestyle{empty}

\begin{abstract}

% Generative Language Models have been trusted into the public eye with the release of GPT-3 and, more recently, with ChatGPT. 

% While it remained a niche research domain until recently, their recent performance has caught public attention and led to both a general excitement over the opportunities they offered, as well as concerns regarding the capabilities of such models, as well as malicious applications.

% Here, we perform a quick review of the current history and SotA of Generative models and their implications for their capabilities, limitations, and future developments.

\noindent
Generative Language Models gained significant attention in late 2022 / early 2023, notably with the introduction of models refined to act consistently with users' expectations of interactions with AI (conversational models). Arguably the focal point of public attention has been such a refinement of the GPT3 model - the ChatGPT and its subsequent integration with auxiliary capabilities, including search as part of Microsoft Bing. Despite extensive prior research invested in their development, their performance and applicability to a range of daily tasks remained unclear and niche. However, their wider utilization without a requirement for technical expertise, made in large part possible through conversational fine-tuning, revealed the extent of their true capabilities in a real-world environment. This has garnered both public excitement for their potential applications and concerns about their capabilities and potential malicious uses. This review aims to provide a brief overview of the history,  state of the art, and implications of Generative Language Models in terms of their principles, abilities, limitations, and future prospects -- especially in the context of cyber-defense, with a focus on the Swiss operational environment.

\end{abstract}

\noindent
{\small
}

\noindent
{\small
\textbf{Highlights:}
\begin{itemize}
    \item Overview of principles, abilities, limitations, threats, and future prospects of attention-based generative language models
    \item Implications for the cyber-defense, specifically within the Swiss operational environment
    \item Threats: information operations, information leakage, better web indexing, phishing, falsification, triangulation, empowering unsophisticated attackers, vulnerabilities injection, and control hijacking
    \item Mitigation: LLM capabilities, training data, and training monitoring, model red-teaming, digital operations good practices (logging and anomaly detection, on-premises deployment, external resources blocking, encryption by default, off-site backup, ...); data smoke-screening, end-user education, differential privacy, formal verification, defense-in-depth
\end{itemize}
}

\noindent
{\small
\\ \textbf{Outline:} This review is organized as follows. Section \ref{sec:Introduction} provides the fundamental principles behind modern soft-attention-based generative models (LLMs) and basic insight into their evolution. Section \ref{sec:GPT} presents the GPT family, while section \ref{sec:BaseLLMs} presents other notable base large language models (LLMs), and Section \ref{sec:ConvoAgents} focuses on alternative conversational agent LLMs. Section \ref{sec:LLMLimitations} provides fundamental limitations of soft attention-based LLMs, while Section \ref{sec:swiss_cyber_defense} focuses on threats implications for Swiss cyber-defense and suggests potential mitigation avenues. Section \ref{sec:Forecasts} attempts to a forecasting analysis related to short-term generative LLMs development and adoption.
}

% \alain{
% Une short « Tech Review » de max. 4-5 pages d’ici mi-février sur le modèle utilisé dans notre livre ci-joint. But : présenter les enjeux des modèles génératifs pour les non-experts avec une focus sur les implications pour la cybersécurité
% }

\section{Introduction}
\label{sec:Introduction}

\subsection{Deep Neural Language Models}

Understanding Deep Neural Language Models (LLMs\footnote{LLM stands for \textit{Large Language Models} and is the commonly used abbreviation for Deep Neural Language model in contemporary literature. As such, we use the same abbreviation here, even if the models we are discussing are not necessarily large by early 2023 standards.}) requires several important departures from an intuitive understanding of natural language. First, from the point of view of LLMs, letters, words, or even sentences do not exist (as defined by human agents). They operate in a continuous "embedding" space, where segments of characters of a fixed length are interpreted as vectors based on how close their meanings are. Elements of such embedding space are often referred to as \textit{tokens}\footnote{The optimal way to convert a text to tokens and back has been subject to extensive research. Finite-length tokens were one of the first approaches. However, modern approaches use varying lengths depending on the context, for instance, for digits.}. Second, from the perspective of LLMs, a suite of such tokens is nothing more than a trajectory, a line, in that continuous "embedding space." Third, such trajectories are not deterministic but rather probabilistic distributions, indicating how frequently trajectories combining a suite of similar tokens in similar order have been encountered in the training data.

While this representation is highly counter-intuitive, it is not exactly new. Introduced in the early 1990s by \cite{ProbabilisticTokenPredictionByNNs1991Mikkulainen} and \cite{ProbabilisticTokenPredictionbyNNs1996Schmidhuber}, shortly after it was popularized by \citet{ProbabilisticLanguageModel2000YBengio}, it was shown to outperform existing state-of-the-art methods in tasks such as machine translation \cite{Schwenk2006}, as long as it was provided sufficient training data, which at the time was made possible by Google Web crawls and digitization of existing translations, such as EU Council parallel texts. However, this model was not only suited for translation. A large number of Natural Language Processing (NLP) tasks could be represented as sequence-to-sequence translation \citep{UnifiedArchitectureForNLP2008CoolobertWston}, including text generation. 

However, this approach had a major problem - learning and representing the trajectories in the "embedding space." Whereas rule-based chatbots could have a combination of pattern matching and response "else-if" rules, LLMs learned by themselves, from massive amounts of data and in high dimensions. A first breakthrough was achieved by using Recurrent Neural Networks (RNNs) \citep{RecurrentNNLanguagemModels2010Mikolov}, and a proper text generation in a simple entailment context\footnote{An example would be "Complete the suite: "Dog, cat, cow, ..." with "goat," "sheep," or other domestic animal being an acceptable answer} has been shown to be possible by \citet{TextGenerationAsTranslation2011SutskeverHinton}, after the introduction of an improved algorithm to train RNNs.

Despite their great initial performance, RNNs suffered from two major problems. First, their training was inherently sequential. The principle of RNNs consists in reusing the output of a processing cell (neuron) as its own input. This is what makes these neural networks recurrent, but it also means that processing cells cannot start processing the next item in a sequence before it is done with the previous one. Second, the length of sequences RNNs is practically limited due to the information about the previously seen sequence contents being passed as a recurrent output that eventually vanishes, as well as the fact that every single token they have seen had to be accounted for in the learning and generation process. Not only did it make them unable to pick up any additional context outside that length window, but in the generative mode, it meant they would often generate a distribution of tokens they never saw in their training data and, in the absence of a learned trajectories distribution, fail to generate a continuation \citep{Huszar2015}.

This latter problem was addressed by the "self-attention" mechanism, introduced by \citet{Attention2015YBengio}. The idea was to leverage the fact that for longer token lengths, the space of trajectories effectively encountered in the linear space is sparse, meaning that instead of having to take into account all of the preceding tokens to perform inference, RNN models could be trained only on a smaller set of "important" tokens, that would be selected by a separate "attention" neural network. Widely successful, this mechanism was rapidly adopted for the Google Translate engine \citep{GoogleTranslate2017JeffDean} and, to the best of our knowledge, is still in use there. The interesting property of that "attention" neural network is that it was not recurrent. It was unnecessary to compute an attention vector for one sequence before computing it on the other, meaning it could be efficiently parallelized. 

\begin{figure}
\centering
\includegraphics[width=0.95\textwidth]{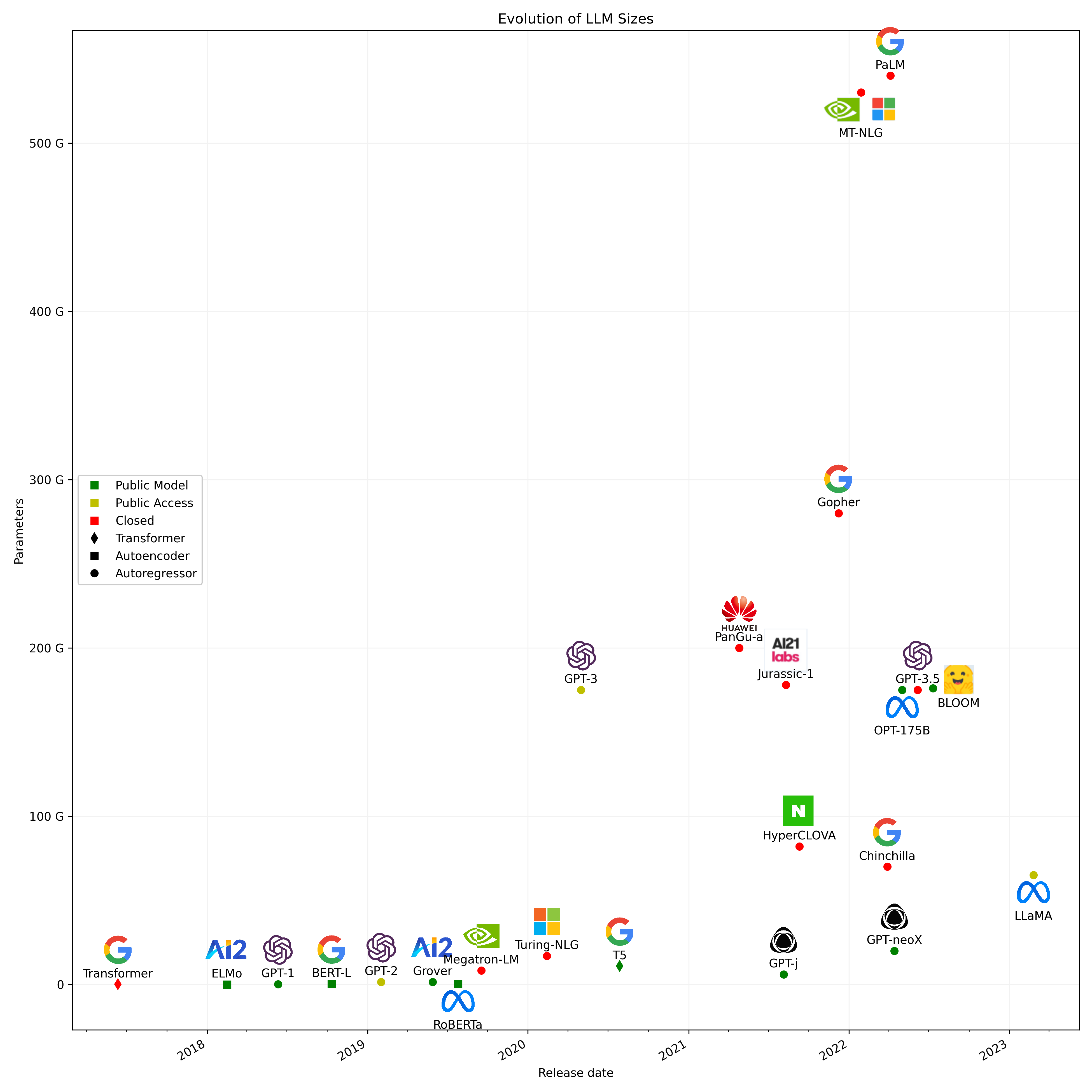}
\caption{\label{fig:HuggingFace} Evolution of size, type, and availability of common LLMs. Due to different model size scaling laws, Mixture-of-Experts (MoE) models have been omitted.}
\end{figure}

The now-seminal "Attention is all you need" by \citet{Transformer2017Google} demonstrated that by increasing the overall size of the model and adopting a multi-layer, multi-head pure attention architecture with normalization between layers (i.e., the Transformer itself), RNN processing units could be removed from the architecture. Without recurrent elements, the network did not need to wait anymore to process prior tokens to obtain the recurrent input for the next token but instead could be trained synchronously. While it might not seem as much, it is hard to understate how transformative it was. Model training can now be fully parallelized across as many computation nodes as was available to the training team, leading to an arms race in model size, training dataset size, and the amount of computational power invested into training them. The resulting exponential growth, represented in Fig.\ref{fig:HuggingFace} 
% \dimitri{This figure is barely readable. Of course, we get the message, but would be of ease to the reader if we increase the size of both axes' labels and logos. I know the figure comes from Hugging Face, but then we will need to fix it by reproducing it?} \andrei{Right now it's a placeholder, and it's also a derived work from the one of HuggingFace. I was going to re-do the graph before the publication anyway to improve the readability}
, was halted only by the depletion of available training data. 
% \dimitri{[thinking out loud:] Does it make sense to speak about the fact that today, the restriction is precisely the amount of training data? (citing some  large models that cannot before as they have too many parameters and not enough data to be trained on). Or perhaps, it is too anecdotal for an introduction?} \andrei{Yes, it makes sense - it is actually a recurrent theme later in the overview and the rational for some statements I make regarding models that are even larger than GPT-3 or the reason for which GPT-3.5 could exist in the first place. I provide citations later, but I can also provide citations here - notably the Chinchilla paper from Google and the predictability and surprise paper from Anthropic.}

\begin{figure}
\centering
\includegraphics[width=0.5\textwidth]{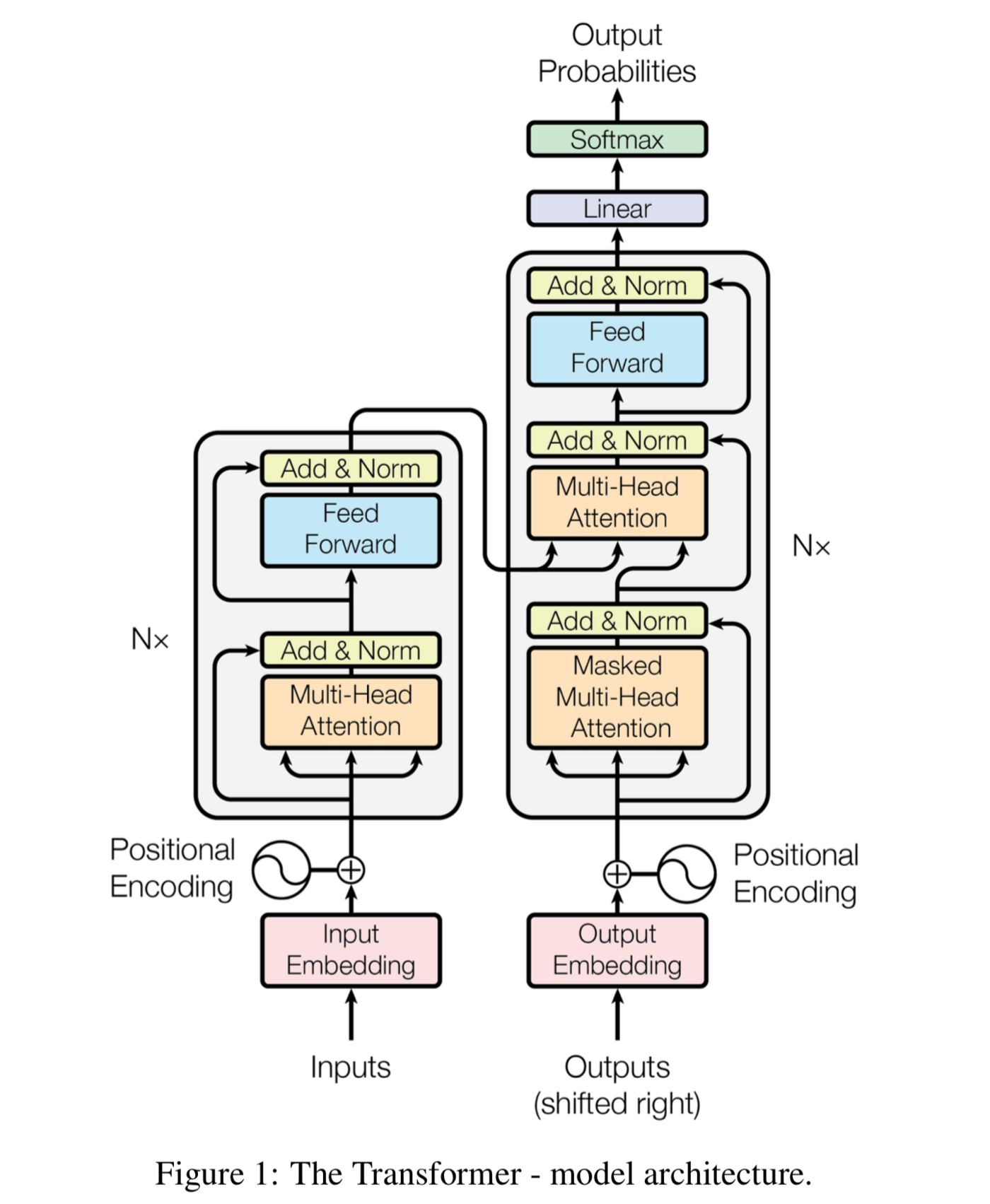}
\caption{\label{fig:TansformerArchitecture} Encoder-Decoder Transformer architecture as presented in \citet{Transformer2017Google}}
\end{figure}

One of the interesting features of the Transformer is that due to its intended use as a neural translation engine, it contains two mostly independent parts. First, the encoder, whose role is to parse the input text and produce an "embedding space" representation of it. Second, the decoder receiving that representation will generate the corresponding translation one word at a time while looking at previously generated words to make sure the next one is still coherent with them (Fig \ref{fig:TansformerArchitecture}). 

As such, models that are not translation-specific can use only the decoder part if they are generating text based on only the previous token and only the encoder part if they don't necessarily seek to generate text but rather understand the structure of the text. Because of that, purely generative text models tend to use only the decoder part, whereas models destined for a more general text understanding tend to use the encoder or both parts of the transformer.

\subsection{Generative Deep Neural Language Models}

From the point of view of a Transformer-based LLM, generating new text is equivalent to the generation of a translation, except that there is not necessarily an embedding space representation. Instead, it's the continuation of an initial word sequence, where the word to be generated can be located either at the end of the original sentence or in the middle of it. These two cases correspond to the two main approaches to training Generative Deep Neural Language Models.

\textbf{Autoregressive} models are trained by a random sampling suite of words in the source dataset ("utterances"), masking a word and training the model to predict the masked word accurately  \citep{Autoencoding2010BengioBrothers}. Autoregressive models are best thought of as autocomplete - they are trained to complete sentences in a way that is representative of their training dataset.

\textbf{Autoencoding} models are trained similarly, except that the masked word can be located anywhere in the randomly sampled suite of words \cite{Autoregressive2016BowmanSBengio}. Autoencoding models are best thought of as search or autocorrect suggestions - they are trained to rewrite whole sentences in a way that is representative of their training dataset. Just as Google search suggestions, they can suggest words at the end of a query to refine it, but they can also add one at the beginning or even rewrite a word (for instance, to correct a typo). While Autoencoding models can be used to generate text based on utterances, their strength is rather in applications that require understanding the utterance as a whole.

While the autoencoding models are generally considered to be more powerful than autoregressive ones, their generative capabilities are not necessarily optimal for the size of the model, training dataset, or the computational resources spent training the model. After all, the training mode relevant to the generation represents only a fraction of the training time of autoregressive models.

There are several paradigms of how generative models can be trained. However, only one is currently dominant and is referred to either as "generative pre-training" or "teacher forcing." Specifically, the model is provided with a large set of utterances with the last word masked and is asked to predict that last token. Based on the proximity of predicted tokens to the tokens in the training dataset, a loss is calculated, and the model is trained through back-propagation.{\footnote{Backpropagation is an algorithm used to train artificial neural networks. The goal of backpropagation is to adjust the weights of the neurons in the network, with the final purpose of minimizing the error between the predicted outputs and the actual outputs.}} Each set of such utterances is commonly referred to as a "batch." In some cases, late in the training process, a refinement process is possible when the model is allowed to predict more and more tokens to stabilize the generation of longer texts \citep{Huszar2015}.

\subsection{Generating text}

Once trained, LLMs can then be used to generate new text. For that, they need a starting text called "prompt," for which they will be looking for the word that would most likely be continuing that prompt in their training dataset. However, as we explained above, models don't learn specific words but rather probabilities of related tokens. Hence, when they are being used to generate texts, for every word, they can only provide the probabilities of \textbf{all} words in their vocabulary. Hence, to actually generate a single next word, they need a "sampling strategy" that would choose that single word.

Four sampling strategies are most used: \textit{maximum likelihood}, \textit{beaming}, \textit{top-K}, and \textit{top-p/nucleus}. The latter is considered to be State-of-the-Art (SotA) and has been introduced by \citet{SamplingStrategies2020Holtzman}, which also reviews other sampling strategies.

Maximum Likelihood always picks the most probable next word. While it can be a good idea for short texts with long prompts, the model is likely to end up in a state where the chosen chain of words has no continuation it would know of, and it would start generating nonsense. This is known as "output degeneration" \citep{Huszar2015, SamplingStrategies2020Holtzman}. Beaming allows us to mitigate some of those issues by looking at the cumulated probability of the next $X$ words and choosing the word that maximizes the probability of having a high-probability branch. 

However, in both cases, the same prompt will generate a similar, if not the same, response, which is usually not what is wanted. For instance, getting always told the same tale in response to a "Tell a tale starting with \textit{Once upon a time}" would be a bit boring, especially since the model is capable of doing better. That is specifically the problem that the top-K sampling is meant to solve. Rather than sampling deterministically, it randomly samples one of the top K most probable tokens. While it solves the repetitiveness problem, it creates the new one - in a setting where a unique suite of words is the only answer, it is susceptible to pick one of the other K continuation words. 

Finally, top-p tries to combine the best of the top worlds by sampling randomly out of the tokens with the highest probability, such that their cumulated probability stays above $p$. In this case, a token that almost surely needs to be generated will be alone in the random sampling pool, whereas in the cases where many different continuations are acceptable, one will be chosen at random.

Once the next token has been generated with one of the sampling strategies, it is added to the original prompt, and that combined text becomes a prompt for the generation of the next token.

\subsection{Filtering out Undesirable Items}

While the basic generative strategies allow the model to generate texts that are similar to what it has seen in the training dataset, with large and heterogeneous datasets, such as results of internet crawls, not all texts are necessarily what the users want to see as output. Classically, things such as instructions for the best ways to commit a crime, slurs, or generally toxic content needs to be filtered out. 

Three approaches exist to solve that problem. 

First, the \textbf{model fine-tuning} will train the model on a dataset of prompts by encouraging desired output and discouraging undesired output through a custom loss function, where the loss is no longer defined by whether the model can predict a continuation to a prompt, but whether the output is toxic or criminal. There is no possibility of knowing in advance how well the fine-tuning worked, and models that have been fine-tuned are generally more prone to output degeneration \citep{T52019Google, AIBorder2022Anthropic}.

Second, the \textbf{guided sampling} will take the newly generated continuations to the prompt by the model and apply independent classification text models (sometimes referred to as "critics") and force the model to re-start generation from a previous prompt state if they detect that the model is generating undesirable output. One of the best-known examples of such critic-guided sampling was proposed by \citet{GeDI2021SalesForce}. 

Third, the \textbf{implicit pre-prompting} will leverage the model's ability to predict possible responses to an utterance - including the ones stating its undesirability. Initially described by \citet{ToxicityPrompts2020McGuffie}, the idea is to modify the way LLM generates text in response to users' prompts by adding additional context for the model through additional prompts before letting an end-user access it. With the demonstration that LLMs were able to correctly classify their own output as biased, toxic, or hateful \citep{SelfDebiasing2021Schick}, pre-prompting LLMs to avoid such behavior became a standard approach \citep{InstructGPT2022OpenAI}. A variation of this approach with sometimes explicit prompts is the chain-of-thoughts prompts, asking the model to reason step-by-step and provide the rationale for the response, in turn leading the model to improve in basic logic and arithmetics tasks \cite{ChainOfThoughtsPrompting2022Google}. 

A variation combining the second and third approaches is the self-guided generation. In a self-guided generation, a separate instance of the model is asked to critique the outputs of the instance interfacing with the user, and that critique is used to censor the outputs of the latter model.

\subsection{Memorization vs Generalization}

LLMs learn the distribution of token sequences in the model embedding space based on the data present in the training set and generate utterances based on a sampling strategy and a prompt. This means that they are consistently on edge between citing the elements of the training dataset they have memorized, if the prompt is sufficiently precise and unique, and composing a new, never-seen continuation to the prompts. The former is referred to as the "memorization" ability of the model, whereas the latter is "generalization." Historically, "memorization" of the models has been thought of as overfitting the training data and easily avoided by exposing the model to the same training as little as possible.

However, results that followed shortly after the first GPT models release - notably \citet{GPT2LeaksDataAF2021} have shown that LLMs are able to memorize things they have seen in the dataset only once, even if a specifically designed prompt needs to be found to trigger the full recall. In this way, authors of \citet{GPT2LeaksDataAF2021} were able to retrieve valid SSIDs, telephone numbers, and emails from the training dataset of the GPT-2 model. Perhaps even more impressively, the GPT-2 model authors were using was able to recite 834 digits of Pi, which it has encountered in the source dataset only once.

However, as of today, there are no known rules or approaches to either improve or discourage memorization for the models, and the research into prompts for retrieving private information is an active domain of research, often referred to as "Model Red Teaming" \citep{RedTeamingLLMs2022}. As a rule of thumb, currently, no data used to train an LLM can be considered safe, and no text generated by an LLM can be assumed to be factually correct or as not containing memorized information.

\subsection{Effect of the Model and Training Dataset Size}
\label{sec:intro_scaling}

% \dimitri{The following sentence is unclear to me:}
% \andrei{How about now?}

The dramatic growth in the models' size between 2019 and 2022, illustrated in Fig.\ref{fig:HuggingFace}, has been driven by almost perfect predictability of the generative models' performance increase. As long as it is provided with a sufficient amount of data and computational resources to train on that data, a larger model is going to keep improving its ability to predict the next word in a text - across a variety of contexts represented in the training dataset \citep{ScalingLawsLLM2020OpenAI, UnpredictabilityOfLLMs2022Anthropic, ComputeOptimalLLMs2022Google}. Correspondingly, that translates to a better ability of a pre-trained LLM to understand a variety of contexts, generate higher quality, more nuanced, and longer texts, and remember more context present in the prior text it is encountering.

% The dramatic growth in the models' size illustrated in Fig.\ref{fig:HuggingFace} is driven in large part by so far perfectly the predictable improvement in the quality of text generated by an LLM depending only on the number of model parameters, training dataset size, and the computational power invested in training the model \citep{UnpredictabilityOfLLMs2022Anthropic}. 

While the predictive base model performance is an interesting ability in itself, after exceeding a certain size, the models start unlocking new capabilities. For instance, between 10 and 100 B parameters, LLMs start being able to generate, compile, and run computer code, translate between languages, or emit predictions of human behavior - at the level of specialized models or better. Such \textbf{emerging capabilities} have so far been impossible to anticipate, and it is not entirely clear what additional capabilities even larger models might acquire \citep{UnpredictabilityOfLLMs2022Anthropic}. 

However, with an increase in size, LLMs not only unlock emerging capabilities but also \textbf{emerging failure modes}. Right around the time that LLMs learn to program and play chess games, they also acquire bias based on sex, gender, race, and religion \citep{UnpredictabilityOfLLMs2022Anthropic}.

Overall, those abilities are generally believed to be made possible through a combination of a bigger attention span of the model, allowing it to take in more context, a larger "hidden state," allowing it to encode more different context-continuation matches, and finally, more parallel layers that allow learning more different ways to map texts to "hidden states."

\begin{figure}
\centering
\includegraphics[width=0.95\textwidth]{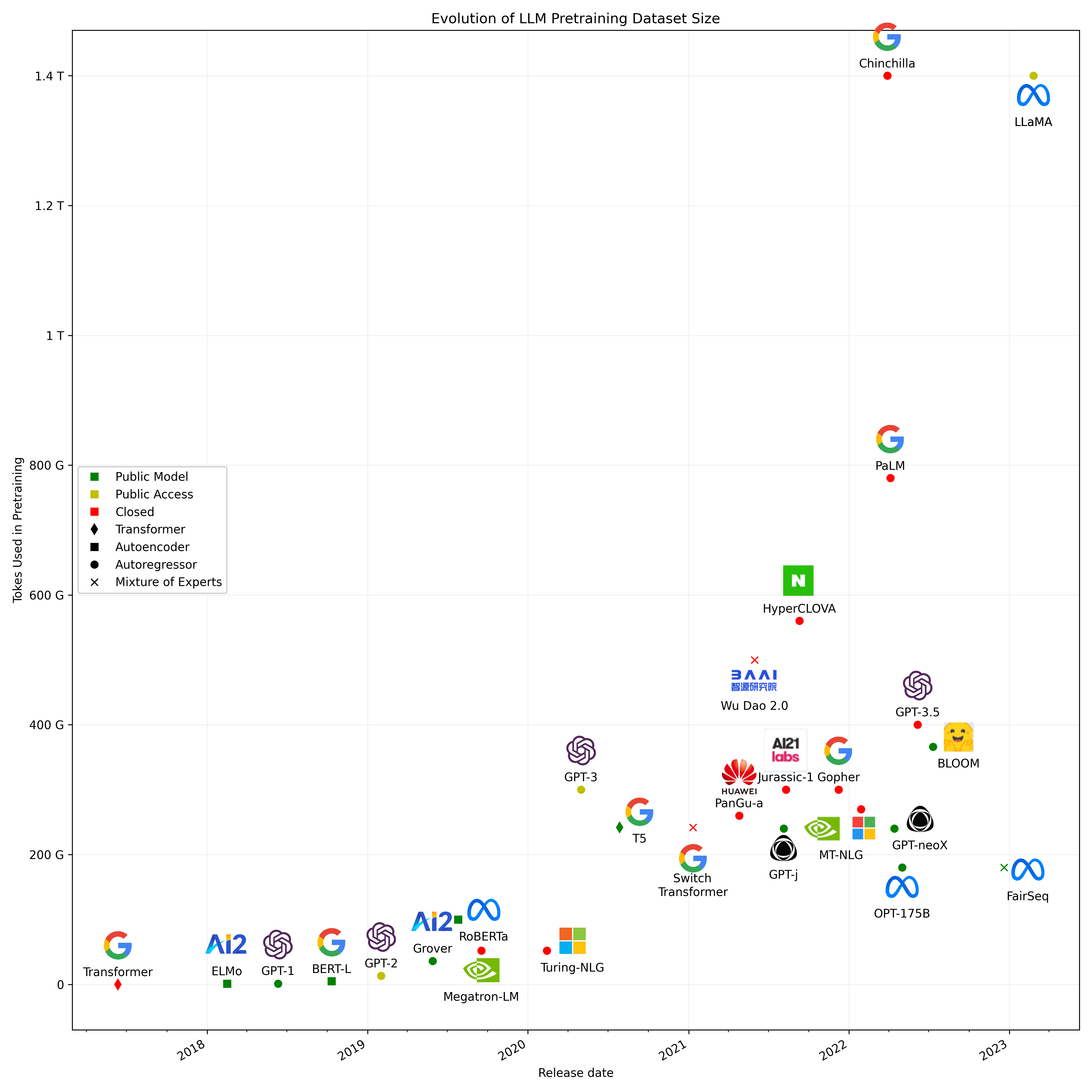}
\caption{\label{fig:training_datset_progress}Progression of the pretraining dataset size for common LLMs over time. The pretraining dataset size in tokens has been taken from the model announcement whenever available and otherwise estimated at 0.3 tokens/byte, based on the results presented by \citet{Bloom2022HuggingFace}\protect\footnotemark.}
\end{figure}

While the abilities of the model are unclear, the consensus seems to have emerged among researchers that there is a relationship between the model size, dataset size, and computational power investment required to achieve optimal resource utilization, that is, approximately 10 tokens/parameter with computational power requirements multiplied by 4 with every model size doubling \citep{ComputeOptimalLLMs2022Google}. When taking into account the training dataset size rather than the model alone, a completely different picture emerges. Fig.\ref{fig:training_datset_progress} shows the progress of the training dataset size for notable released models, suggesting an explanation as to why some smaller models are known to perform particularly well compared to the models of similar or even larger size (RoBERTa, T5, GPT-j, GPT-neo(X), GPT3). In Fig.\ref{fig:honestsizes}, we attempted to renormalize notable models to the training dataset size, if they were trained according to the optimal resource utilization rule. We discuss this renormalization and its limitations in section \ref{LLMs_aint_getting_bigger}.

\footnotetext{Based on its API description GPT3.5 was assumed to have been further pretrained on the same data as the CODEX model, with the dataset size extrapolated according to the programming language distribution in the data in \citet{Bloom2022HuggingFace}.}

\section{GPT family}
\label{sec:GPT}

Generatively Pre-trained Transformer (GPT) LLMs are a family of autoregressive generative models based on only the decoder part of the original Transformer and ranging in size from 117M (GPT-1) to 175B parameters (GPT3-XL "GPT3") developed by OpenAI between 2018 and 2020. 

With the GPT3.5 release in 2022, OpenAI ceased providing detailed documentation of its models' architecture, training, and capabilities, a tendency that was further confirmed with the staggered release of BingChat/GPT4. in February-March 2023. While we treat BingChat/GPT4 as parts of the GPT family here, the nature of the model is likely to be radically different from the previous GPT family generations and should be treated separately. We discuss this in section \ref{subsec:binggpt}.

\subsection{GPT1}

The first model in the family, GPT-1 \citep{GPT12018RadfordSutskeverOpenAI}, directly re-used the decoder of the base Transformer \citep{Transformer2017Google} architecture, using 12-layers, 12-heads/layer, 768 hidden states per head, and an attention span of 512 tokens, trained on the BookCorpus \citep{BookCorpus2015} dataset of unpublished books, containing around 1B words. The BookCorpus was chosen to achieve a long-distance coherence 

\subsection{GPT2}

The GPT-2 generation came in four sizes, 117M, 345M, 762M, and finally, 1.5B parameters, essentially preserving the original architecture \citep{GPT22019RadfordSutskeverOpenAI}. The modifications mostly focused on scaling up the model by increasing the number of layers (12 to 48), the number of hidden states per attention head (768 to 1600), and finally, the size of the context taken into account by increasing the attention span. In addition to that, the model's architecture and training have been modified to make it more stable, by modifying the normalization structure and increasing the batch size. Finally, to support training such a large model, a new training dataset has been compiled by authors, combining the text contained at all the outgoing links of a popular social media website Reddit, as a proxy for the text being of sufficient quality to be of interest to human readers. The obtained dataset, at around 10B words, was used to train all the GPT-2 models, and the largest GPT-2 model was withheld by OpenAI to prevent malicious use.

\subsection{GPT3}

The GPT-3 generation was an immediate successor to the GPT-2 one and included 8 pretrained models of different sizes, ranging from 125M to 175B parameters \cite{GPT32020BrownSutskeverOpenAI}. This time around, the increase in size also involved the number of attention heads in the model, as well as the number of layers, the number of hidden states, and the length of the attention span. The largest model - GPT-3 175B "GPT-3" had 96 layers with 96 attention heads per layer, 12200 hidden states, and a context of 2048 tokens. To train this model, authors used a second iteration of the dataset used to train GPT-2, along with a filtered and deduplicated version of the Common Crawl, a dataset of all the text that can be found by a crawl on the Internet, as well as two large datasets of books and Wikipedia texts, for a total of approximately 400B words. However, to account for the quality of texts in each database, the actual training dataset is generated by weighted sampling from datasets. For instance, a token from Wikipedia is likely to be seen on average 3.4 times during the training compared to 0.44 times for a token from the Common Crawl\footnote{For clarity to an unexperienced reader, we are using "word" and "token" interchangeably, but the two tend to differ, and in fact do differ for the \href{https://platform.openai.com/tokenizer}{GPT tokenizer}, with a token being about 3/4th of a word.}.

\begin{figure}
\centering
\includegraphics[width=0.5\textwidth]{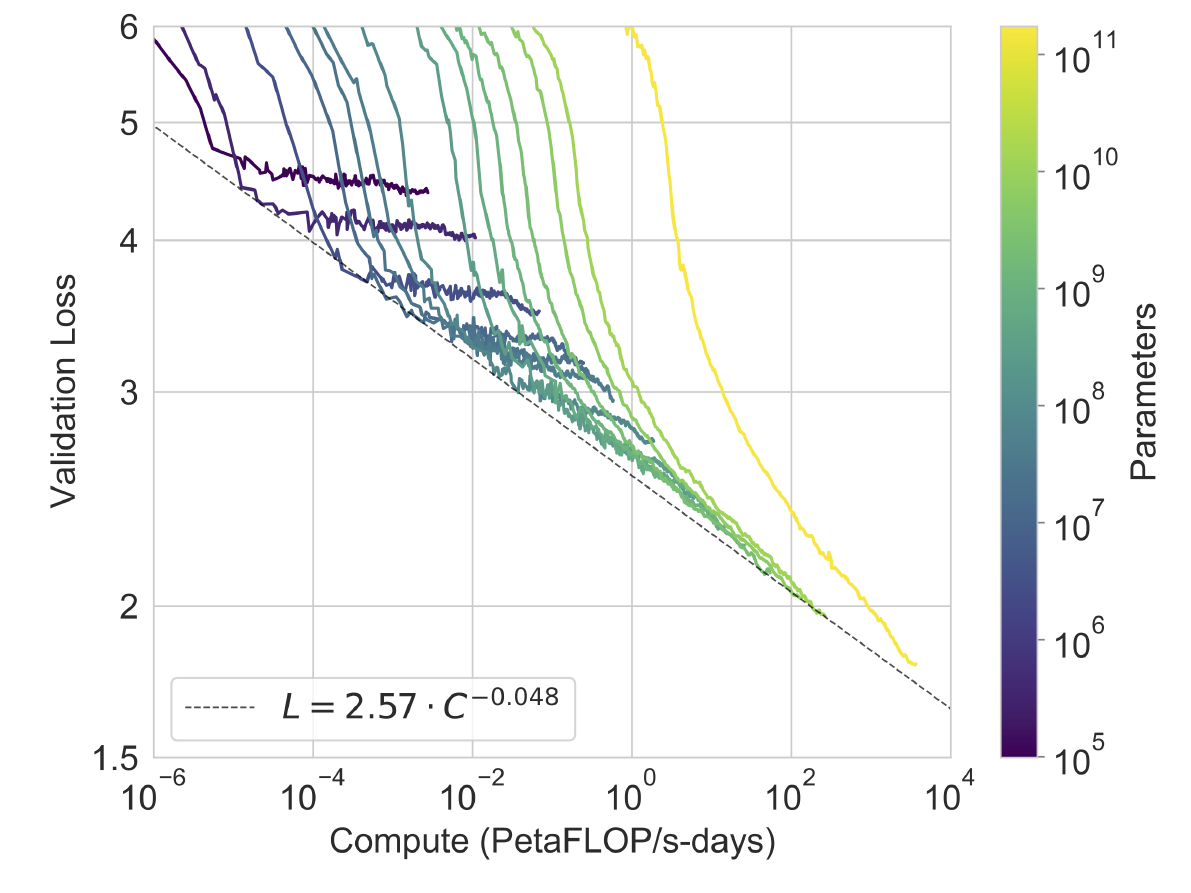}
\caption{\label{fig:GPT3Compute}Relationship between computational power invested to achieve a target single token prediction loss, depending on the size of the model. Figure courtesy of \citet{GPT32020BrownSutskeverOpenAI}.}
\end{figure}

While its performance came with a computational power expense overhead (Fig.\ref{fig:GPT3Compute}), the performance of GPT-3 in the prediction of the next word, given the context, was formidable. 

Despite that, the datasets obtained through crawling the internet at large often contain texts that are generally considered undesirable - ranging from neo-nazi forums to erotic fan-fiction \cite{CommonCrawlisToxic2021, StochasticParrots2021BenderGebruMitchell}. Such texts often start with unremarkable text - for instance, a news item or a mundane scene involving popular characters - which would match common and otherwise unremarkable prompts. However, the inappropriate texts in the training dataset switch from unremarkable to highly toxic and inappropriate. During the training, the LLM learns that such continuations are probable and, during the generation, would sometimes respond to unremarkable prompts with highly undesirable generated content. While this issue was already affecting the smaller models in the GPT-2 family \citep{InappropriateGPT2texts2020PengRiedl}, the larger and more diverse training dataset sourcing for GPT3 has likely significantly worsened the problem and led to more diverse and hard-to-diagnose failure modes.

Similarly, a longer attention span led to the model picking up on correlations representative of past and prior biases that lead to biases in narration and decision suggestion. For instance, a  model could have learned the correlation between higher education and negative life experiences for women in the late 1800s and early 1900s based on their biographies in the training dataset. This is valuable information that is required for the accurate generation on historic themes. However, without mitigation, this correlation can and will also contribute to the model recommending women not to pursue higher education today \citep{StochasticParrots2021BenderGebruMitchell, GenderBiasInEmbedding2019}.

Finally, the interaction with base generative LLMs was highly counter-intuitive for non-expert users. Where the users were expecting an answer to a multiple-choice question after asking one, a generative model would detect a similarity to multiple-choice question collections in their training set and start generating continuations typical in such collections - in other terms, other multiple-choice questions.

To address these issues, OpenAI has opted to try to refine existing model families through a combination of model fine-tuning and guided sampling.

\subsection{InstructGPT}

\begin{figure}
\centering
\includegraphics[width=0.85\textwidth]{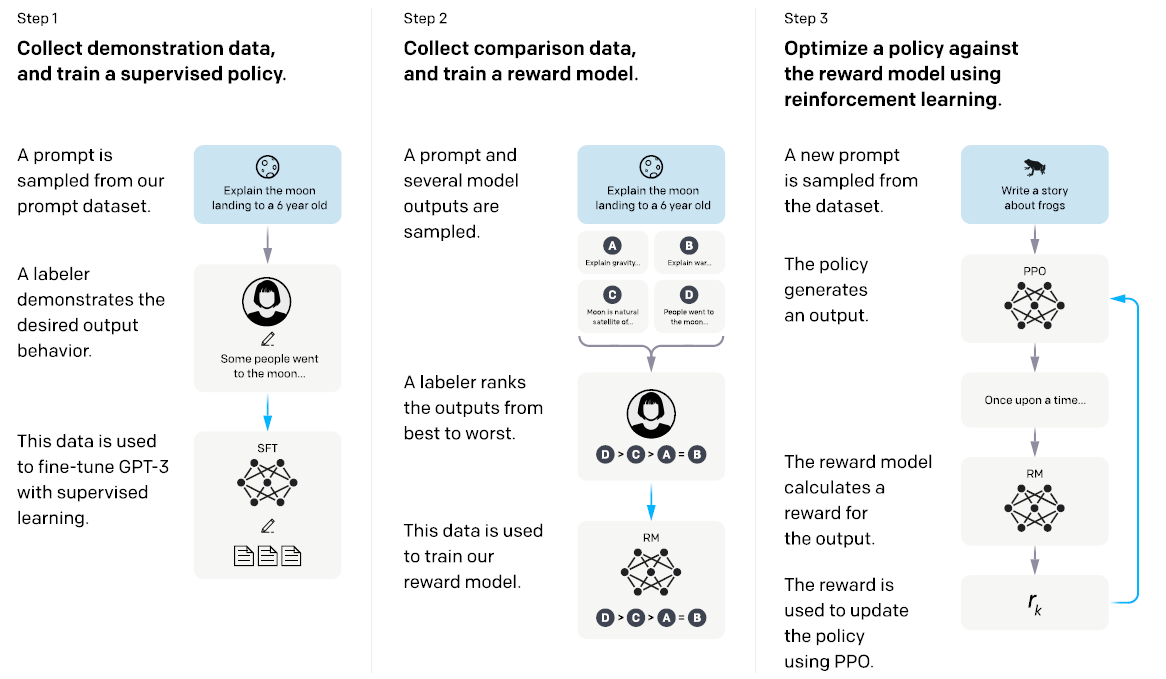}
\caption{\label{fig:InstructLoop} Iterative refinement of an
InstructGPT conversational agents to comply with user expectations. Image courtesy of \citet{InstructGPT2022OpenAI}.}
\end{figure}

InstructGPT takes a member of the GPT-3 pretrained models generation and first fine-tunes the model to respond to "instruction" prompts in a way similar to the one human writers would \citep{InstructGPT2022OpenAI}. This allows the model to answer questions by a human user rather than force a human user to come up with prompts that would lead the model to generate the type of text they desire. As a second phase, human workers rank the quality of the fine-tuned model output along a number of evaluation metrics, ranging from following the constraints specified in the question, to toxicity, to bias to factuality. Their feedback is used to train a "censor" model that is used to guide text generation and further fine-tune the model. Such a secondary fine-tuning is usually referred to as \textit{Reinforcement from Human Feedback} (RFHF). The original InstructGPT-3.6B model has been generally better rated in interactions than the GPT-3 175B models it was compared with. The whole process is summarized in Fig.\ref{fig:InstructLoop}, taken directly from \citet{InstructGPT2022OpenAI}.

\subsection{CODEX}

An interesting immediate application of powerful generative models seemed to be in the automation of code generation. In fact, most programming in corporate environments consists in transforming natural language specifications into a code that passes specifications, often defined as tests (so-called "unit tests").  Interestingly, the largest models in the GPT3 generation already had some capabilities for solving similar tasks due to the presence of code with specifications in their training dataset \citep{UnpredictabilityOfLLMs2022Anthropic}.

With a specific focus on code generation in mind, OpenAI fine-tuned a set of GPT3 models on a large sample of Python code samples from GitHub, PyPI python package manager, and several other sources, all of which contained both a specification and an implementation of the specification. To facilitate their work, the OpenAI team used doctext as a specification, given their ubiquity in Python. Part of the rationale for the choice of Python, besides abundant documentation given its open nature, is the fact that it is one of the most widely used programming languages with a thriving open-source projects ecosystem and is the closest to the English language in its structure among all the major programming languages.

This resulted in a range of models, the biggest of which - CODEX-12B parameters - could solve 72\% of new, human-created coding problems after 100 attempts, and 28\% on the first attempt \citep{CODEX2021OpenAI}. A variant of the CODEX model trained on more programming languages and more code hosted on GitHub is powering GitHub's Copilot. 

\subsection{GPT3.5}

The scaling of the GPT-3 model to the 175B parameters has been determined as optimal given the training dataset size in \citet{ScalingLawsLLM2020OpenAI}. However, subsequent research has shown that the initial experiment by the OpenAI team did not take into account a sufficient variety of architectures, initializing conditions, and model sizes \cite{ComputeOptimalLLMs2022Google}. As such, GPT-3 is now considered undertrained, and in early 2022  OpenAI team further pre-trained those models with additional text data (source unspecified), as well as data used to train the CODEX models, providing it with enhanced code-generating abilities. Based on the code generation abilities of conversational agents based on GPT3.5, we speculate that the dataset was a combination of code snippets with descriptions and annotations from multiple programming languages.

\subsection{ChatGPT}

ChatGPT is the more powerful variant of Instruct GPT, based on the GPT-3.5-175B model. It likely underwent a more extensive fine-tuning and "censor" model training before public release, although the exact information regarding those processes has not been made public.  It also seems that additional "critic" large language models were used to perform prompt filtering to intercept prompts that would lead to the generation of instructions considered harmful (e.g., instructions to make explosives, self-harm, controlled substances, criminal activities, ...). However, again, detailed information on the topic has not been made available by OpenAI.

\subsection{Bing Chat/GPT4}\label{subsec:binggpt}

\subsubsection{BingGPT, Bing Chat, and New Bing}

In early February 2023, Microsoft announced the integration of a successor to ChatGPT with the Microsoft search engine Bing \cite{BingChatAnnouncement2023}. However, to the best of the information available to us at the moment of that announcement, the conversational agent LLM powering the Bing-integrated search-chat has a substantially different set of abilities and behaviors compared to ChatGPT and hence warrants treatment as a separate GPT family generation\footnote{As of the writing of this section, the information available regarding BingGPT was minimal, just as the access to its interface. Hence this part contains a substantial amount of speculation that has been partially confirmed  by the GPT-4 technical report we cover in the next subsection. We opted in favor of keeping this section, given that some of the subjects here are absent from the GPT-4 technical report}. 

The major departure of BingGPT from prior models is that it gains access to auxiliary capabilities. Whereas prior members of the GPT family were purely autoregressive models, whose generation depended on the training dataset, eventual fine-tuning, and prompts alone, BingGPT is able to transform natural language queries to auxiliary services queries (notably search requests) and convert auxiliary services responses back to conversational format, along with references. At the moment of the release, the only known model with similar capabilities is Google's Sparrow \citep{Sparrow2022Google}, based on the Chinchilla LLM \cite{ComputeOptimalLLMs2022Google}, which we cover further in section \ref{sec:sparrow}.

Based on some public demos\cite{LTT_BingChatDemo_2023}, in addition to being to perform search queries, BingGPT seems to be capable as well of: 

\begin{itemize}
  \item Perform image-to-text conversion (image object type, color, logo nature)
  \item Perform basic logic reasoning to split queries (bags of type X that will fit in a trunk of a car Y $\rightarrow$ size of bags of type X, size of car Y trunk)
  \item Perform basic logic reasoning to aggregate information acquired from separate queries (bags size along dimensions vs. trunk size along dimensions; similarity of bags sizes to objects for which there is a record of being put into trunk)
  \item Identification and summarization of customer feedback in a qualitative manner (recurrent points of dissatisfaction or satisfaction rather than a sentiment or a star rating alone)
  \item Requesting further refinement in case of queries allowing for multiple interpretations
  \item Explicitly identifying misspelled but semantically similar search terms, correcting and asking the user to clarify in case of ambiguity (Biomass $\rightarrow$ Bonemass; a videogame boss rather than a fuel or a mass of living organisms)
  \item Offering realistic, search-based scenarios for possible future outcomes regarding a specific domain, technology, or fiction franchise
  \item Potentially, parsing and interpreting sound and visuals of videos to provide a summary and integrate such a summary in a query response results.
\end{itemize}

We speculate that this has been achieved through a combination of fine-tuning, implicit pre-prompting, and choiring\footnote{We refer to a model choir or model choiring architectures that use multiple instances of the same model to delegate tasks such as intermediate summarization, output appropriateness/factuality evaluation, or sub-task extraction and delegation to other LLMs.} of a GPT family base model, specifically:

\begin{itemize}
  \item Fine-tuning and implicitly prompting the model to emit queries to auxiliary search engines or other tools, and forward results returned by them to auxiliary models for summarization and relevance re-ranking, to finally combine the summary of summaries of most relevant results
  \item Fine-tuning with synthetic math and logic problems, combined with implicit pre-prompting with chain-of thoughts prompts to trigger detailed and more likely correct reasoning  and calculations
  \item Fine-tuning and implicit pre-prompting to use an auxiliary pre-trained image-attention-text model such as OpeAI DALL-E \citep{DALLE2021Sutskever}, to trigger image analysis and interpretation in the context of a query
  \item Fine-tuning and implicit pre-prompting to use auxiliary pre-trained voice-attention-text models such as OpenAI Whisper \citep{Whisper2022Sutskever}, to trigger voice analysis and interpretation  in the context of the query, potentially combined with image analysis for videos to enable capabilities described above
  \item Choir prompting, with separate model instances charged with re-formulating prompts, creating intermediate step prompts, evaluating search results relevance or base model output for correctness and alignment.
\end{itemize}

Perhaps the most critical difference a conversational agent would have compared to a traditional search engine, such as Google Search, is the ability of users to provide feedback. In the best scenario, it could allow crowd-sourcing an almost immediate refinement of search results based on the current context, common search mistakes, or spurious correlations, which are known to plague traditional search engines to the point of having interfered with conversational agents' design \citep{SeeKeR2022Facebook, Sparrow2022Google}. In the worst-case scenario, it would allow malicious agents to vector search for their own benefit, either as a part of influence operations or for cyber-criminal economic interests.

While some alignment problems have been reported for BingGPT \cite{BingOffRails2023Verge}, they can potentially be addressed with the data obtained during the open testing of ChatGPT as well as early user experience and feedback for BingGPT itself. Such rectification is, however, far from certain. Some reports indicate that models can be tuned either for safe interactions or helpful interactions, but not both, with a Pareto frontier for a trade-off between the two \citep{AIBorder2022Anthropic}. It is, however, not entirely clear yet if and how it depends on the architecture of the LLM and the data used to train it, so this question remains open for BingGPT.

\subsubsection{GPT4}
A follow-up joint announcement by OpenAI and Microsoft revealed that Bing Chat mentioned above was indeed a novel LLM architecture; specifically, the GPT4 \cite{GPT42023OpenAI, BingIsGPT42023}. Unfortunately, the GPT-4 technical paper \citep{GPT42023OpenAI} lacks almost all of the details necessary for the understanding of underlying architectures, although confirming educated guesses presented above and allowing some new insights.

Specifically for the model size and training dataset, given the scaling of GPT-4 presented in Fig.1, as well as claims that the observed scaling laws were the same as in \citet{ScalingLawsLLM2020OpenAI}, assuming that the next token prediction loss for code and language are comparable, suggests a model size of the order of magnitude of \~17T parameters. Such a model size, with the stated scaling laws, would have required \~28T tokens to train, or about 60x the amount that was available at the time of GPT-3 training \citep{GPT32020BrownSutskeverOpenAI}. Given that the GPT-3 training dataset included the largest clean subset of CommonCrawl OpenAI could use, in addition to custom datasets, and that the largest previously reported dataset stopped at 1.4T tokens \cite{ComputeOptimalLLMs2022Google, PaLM2022GoogleJeffDean}, the origin of 20x the amount of training data compared to what is available to the closest competitor is unclear.

Based on this factor, along with the fact that it uses auxiliary capabilities and both the technical paper and user experience suggest model choiring\footnote{Notably F4 - Similar Chemical Compound purchasing in \citet{GPT42023OpenAI}}, we speculate that instead, GPT-4 is closer in operation mode to Mixture-of-Experts (MoE) models, such as Google Switch Transformer \cite{SwitchTransformer2021Google}. MoE scale differently, with Google Switch Transformer training 1.6T parameters with as little as 200B tokens, suggesting a more realistic 2T tokens for GPT-4 using the same scaling rule, suggesting an underlying base model in the 1.1T parameters range. 

Additional information in the paper so far confirms the speculations we presented above regarding Bing Chat structure and capabilities, with the exception of the video analysis capabilities. Such capabilities have been consistently reported for BingGPT but have not been mentioned in the GPT-4 technical report.

We cover several aspects concerning cyber-security and cyber-defense covered in the GPT-4 technical report in the section dedicated to cyber-security implications. We opted not to cover Microsoft Office 365 Copilot, given a lack of any structured information regarding the underlying model, except for its public announcement \citep{MSOfficeCopilot2023}.

\section{Other Base LLMs}
\label{sec:BaseLLMs}

%\ljiljana{add global introduction to the chapter}
While perhaps the most prominent models among the LLMs, the GPT family is far from being alone. In this section, we focus on base LLMs. Among the models based on the decoder part of the Transformer, just like the GPT family, we distinguish those replicating the GPT architecture as-is and those adapting the architecture in an effort to leverage certain specificities such as  the multilingual aspect (BLOOM) or covering a specific domain (Galactica). BERT and its refinements (RoBERTa, DistilBERT, to name a few), contrarily to GPT, make use of the encoder part of the Transformer and are also suited for tasks other than generations. Finally, Sequence-to-Sequence models, relying on the full Transformer architecture, such as the T5 family, are best suitable for text-to-text transformation tasks such as summarization, translation, but also question answering, and code generation. While differing architectures can make models more or less suitable for some tasks, each architecture can be used for any task, with its success being determined more by model size, training dataset, and training regiment. Notably, all LLMs covered here can and have been used for text generation.

\subsection{GPT clones}
The success of the GPT families led a number of other companies to try emulating their performance by replicating, to the best of their ability, the GPT family. However, given the importance of the role played by the training dataset, in its absence, the clones' performance differs from the base GPT models, and they cannot necessarily be assumed to be interchangeable.

\subsubsection{EleutherAI's GPT-neo, GPT-J, and GPT-neoX}

Developed by EleutherAI, a non-profit collective of NLP researchers, GPT-neo-2.7B, GPT-j-6B, and GPT-neoX-20B \citep{gptNeo2021EleutherAI, GPTJ6B2021EleutherAI, GPTNeoX20B2022EleutherAI} are architectural clones of OpenAI's GPT family at 1.3B, 6, and 20B parameters respectively, with minor architectural variations. The biggest difficulty was replicating the OpenAI training dataset collection and preparation. To replace it, EleutherAI leveraged the Pile dataset \cite{ThePile2021}, a collection of 22 high-quality datasets contributed by varying entities containing about 800G of text, or 240G tokens. 

Despite its smaller size and less pre-processed data, all of the elements of that family are considered to perform well compared to other models. In particular, GPT-J-6B has been successfully used to impersonate multiple human users in a fully autonomous fashion, on a forum-like website, in a real-world setting \footnote{Given the absence of ethics approval, user consent and exposure of users without opt-out to highly toxic LLM output, we consider that specific experiment unethical and will not be citing it here. More information can be obtained in secondary sources covering the incident, notably \citet{GPT4chan2023Verge}.}. Given the model size relative to the training dataset, EleutherAI GPT clone families are likely to be appropriately trained with regards to the findings of \cite{ComputeOptimalLLMs2022Google}.

\subsubsection{HyperCLOVA}

A copy of the GPT family but scaled to 82B parameters and specific to the Korean language, HyperCLOVA was developed by a South Korean Google equivalent, NAVER, and was announced in late 2021 \cite{HyperCLOVA2021Naver}. The main change this model brought was a modification of the tokenizer to better suit the Korean language, as well as a reduction in the model size compared to GPT3, accompanied by the training dataset increase (300B>540B tokens). Interestingly, this modification of the model scaling closer to the updated scaling laws, not yet published at the time, gives further credibility to results in \citet{ComputeOptimalLLMs2022Google}. Similarly, a tokenizer modification suggests that there are potential gains to be made in multi-lingual models by using tokenizers better suited for multiple languages rather than English alone.

\subsubsection{Meta (Facebook's) OPT family}

Following the public attention to GPT-3 on its release, Facebook started its effort to replicate the entire family and, by mid-2022, released the OPT family, a clone of the GPT-3 family, but based on their own training data \citep{OPT2022Facebook}. What is notable is that all models of this family, up to the 175B parameter OPT-175B have been made publicly available. Once again, due to the difference in the data collection and preparation, the performance of the model is generally believed not to match GPT-3 (in large part due to a smaller and less curated training dataset), and the BLOOM team showed it underperformed compared to their own model across all model sizes. This, once again, can potentially be explained by the fact that the OPT-175B paper reported using a significantly smaller dataset than comparable models - at 200G tokens \cite{OPT2022Facebook}, or about 66\% of the dataset size used to train GPT-3.

\subsection{GPT-Like Models}

\subsubsection{BLOOM Family}
Following a foray into minimizing the size of the model with DistilBERT \citep{DistilBERT2019HuggingFace}, in 2022 HuggingFace's research team attempted to explore larger models by partnering with a larger consortium of researchers - BigScience Workshop. BLOOM family of models is the result of that partnership \citep{Bloom2022HuggingFace}. Just like the GPT family, BLOOM is based on the decoder side of Transformer architecture and, for the same size, has fewer layers and more attention heads per layer, as well as a higher number of hidden dimensions. For ~175B parameters, GPT3 is 96 layers with 96 attention heads each and 12k hidden dimensions, whereas BLOOM is 70 layers with 122 attention heads each and 14.3k hidden dimensions. 

Part of this change is justified by the focus of the BLOOM model on increasing multilingual encoding capabilities, maximizing multilingual training data in the original training run, as well as adding over a dozen programming languages into the mix. Additional attention heads per layer are believed to enable parallel encoding of tokens from different languages to the same underlying meaning and sentence structure.

Given the increased focus of BLOOM on multilingual abilities in their model, they invested more effort in compiling datasets representative of languages other than English. In particular, they improved the representation of low-resource languages in the training dataset and extended programming language-specific repositories datasets to include more recent and minor programming languages such as Scala and Rust. Despite that, the model is likely to be focused on the Latin language groups, notably French, Spanish, and Portuguese, and lacks representation of Germanic languages outside English.

The authors demonstrate their model outperforms the OPT family across a range of tasks at all model sizes and is comparable to the GPT family when it comes to bias and toxicity. Unfortunately, no third-party evaluations of the model are currently available. However, given HuggingFace's central role in collecting and distributing pre-trained language models, we believe that the BLOOM family could become the open-source standard for generative LLMs in the 10-200B parameter range.

\subsubsection{Compute-Optimal Models}
\label{sec:computeopt}

Alongside the release of their GPT-3 model, OpenAI published a report detailing the scaling of large language models performance with its size, justifying the choice of the GPT-3 model size given the data they had access to at the time \citep{ScalingLawsLLM2020OpenAI}. However, as we mentioned previously (section \ref{sec:intro_scaling}), these scaling laws have since been shown to underestimate the number of tokens needed to train the model to optimality \citep{UnpredictabilityOfLLMs2022Anthropic, ComputeOptimalLLMs2022Google}. 

Based on these new scaling results, a new generation of LLMs has been developed and trained. While smaller in size than the GPT family, such LLMs have been shown to match and even exceed the capabilities of GPT family models 3x their size. The three most visible generative autoregressive models in this category are Google's Chinchilla \cite{ComputeOptimalLLMs2022Google}, Facebook/Meta's LLaMA \cite{LLaMA2023Facebook} and Anthropic's base LLM for Assistant and Claude, alluded to in \cite{AssistantLLM2022Anthropic, AIBorder2022Anthropic}. Ranging in size between 52B parameters for Anthropic's base LLM and 70B parameters for Chinchilla, they all compare favorably to GPT3-175B while all being easier to deploy and run thanks to a smaller size.

As a side note, RoBERTa \citep{RoBERTa2019Facebook} and T5 \citep{T52019Google} models can also be argued to be compute-optimal as well. While they are not autoregressive and have not been designed for pure text generation, they have generative capabilities, and T5 is commonly used in this role. However, what makes them compute-optimal is the fact that they satisfy the empirical compute-optimal scaling rule of 10:1 between the training dataset size in tokens and the number of model parameters. Finally, for the same reason, GPT-J and GPT-neo models from EleutherAI can be considered as potentially compute-optimal.

Of particular concern for cyber-defense is Facebook/Meta's LLaMA model, whose 13.5B parameter variant has been claimed to match 175B GPT-3 model performance while fitting in the memory of single consumer-grade graphics cards. While the technical paper accompanying the model raises some questions - notably with regards to LLM models scaling and the training dataset used to train it\footnote{Authors claim to have used two datasets as largest sources of training data that are considered as redundant - namely the Common Crawl and Colossal Cleaned Common Crawl (C4). Researchers who created the C4 dataset provided experimental evidence that their dataset was strictly superior to Common Crawl when it came to training LLMs \citep{T52019Google} and should be used instead of the whole Common Crawl whenever possible.}, the LLaMA-13.5B model remains a powerful SotA generative LLM that can be fine-tuned for downstream applications. 

What makes this model so concerning is the fact that its weights were leaked on the 4chan message board, a community known for its adjacency to cyber-criminal circles and extensive usage of ad-hoc information operations tactics (raids, harassment campaigns, de-platforming through mass reporting, ...). Unfortunately, that community has an excellent idea of how LLMs can be leveraged for such goals, given that they were exposed to disguised LLMs first-hand in mid-2022. At this point, a Swiss ML influencer flooded a 4chan board with output from highly toxic conversationally fine-tuned LLMs as a part of an experiment evaluating LLM detectability without obtaining any ethical approval, subject consent, or having a harm mitigation plan in place \cite{GPT4chan2023Verge}. As such, users of 4chan now have resources, technical knowledge, and a practical understanding of the strengths and limitations of using LLMs for impersonation.

\subsubsection{Larger and Domain-Specific Models}

When OpenAI pushed the model sizes further than anyone before with GPT-3, they were not alone. Several other companies pursued larger LLMs, leveraging the parallelism of pure self-attention architectures. In 2021 Google unveiled the 280B-parameter Gopher language model \citep{Gopher2021Google}, whereas NVIDIA partnered with Microsoft to develop a 530B Megatron-Turing NLG model, announcing it in 2020 \citep{MegatronLM2019Nvidia}. Beijing Academy of Artificial Intelligence announced having trained a 1.75T parameters model in 2021. However, no accompanying papers have been published, and the model has not been made accessible to third parties for validation. 

However, as we mentioned before, GPT-3 was already using most of the written text publicly available on the internet and, in the end, is thought to be undertrained for its size \cite{ComputeOptimalLLMs2022Google}. Due to the training dataset limitations, larger models did not offer further performance improvement and were abandoned, at least until now.

Similarly, for reasons of data availability, attempts to build domain-specific LLMs have met little success. A prominent example of such failure is Meta's (Facebook's) Galactica\footnote{While Galactica has been conversationally fine-tuned and, as such, is a conversational agent, it is much closer to other LLMs in capabilities and shortcomings. Hence we decided to still treat it here as a base LLM.}, trained on the dataset of texts representing scientific knowledge \citep{Galactica2022Facebook}. Given the mismatch between the size of the training dataset (~100B tokens), sparsity of information in each domain (aka low coverage of the same facts with varying formulations), and comparatively large model size (~100B parameters), the model was unable to learn inference robustly. Combined with the inherent difficulty of soft-attention architectures to distinguish truth from falsehoods, the resulting model would easily start generating counterfactual or even nonsensical texts, despite training on scientific and educational texts \citep{GalacticaPull2022ArsTechnica}.

Finally, in the space of large models, in 2022, Google released its Pathways Language Model, PaLM \citep{PaLM2022GoogleJeffDean}. PaLM leverages the newly introduced Pathways scheduler \citep{Pathways2022GoogleJeffDean} to train a range of models, from 8B to 540B parameters, while injecting a substantial amount of text data from social media interactions. Approximately doubling the GPT3 training data, it is still unclear whether the model has enough training data to achieve proper performance. Despite achieving SotA and, finally, matching average human performance on a panel of 58 tasks, the results indicated by authors suggest that PaLM architecture compares unfavorably to the Chinchilla model at 70B parameters \cite{ComputeOptimalLLMs2022Google}, with PaLM significantly underperforming compared to Chinchilla at similar model sizes, and offering only a marginal improvement at the cost of 7x increase in size.

% Breaking with the general-purpose pretraining popularized by the GPT family and BERT-like models, PaLM combines 

% going in the opposite direction compared to OpenAI's general-purpose language models. Rather than trying to train a single model, they train a range of LLMs for specialized tasks and combine it into a single super-model punctuated by non-differentiable binary switches, with switch decisions being learned through Evolutionary Algorithms. According to them, this approach would allow switch-over that could not be bypassed between specific functions, which would be critical in such applications as search. A specialized LMM would parse user query, then pass it to a set of networks to extract requests for factual information (date, name, event, ...), which would, in turn, pass it to a network generating a database query with proper fields and formatting, which would, in turn, pass the factual information to other networks to wrap it in response and add context, if needed. While the premise sounds promising, the PaLM has been in development since 2017, and at 580B parameters, it is not entirely clear when its training might converge

% A more lightweight approach to enforce factual correctness has been presented by <citet>, where a language model is fused with a database of known events, and a critic detecting event statements in the generated text will check them in a factual database and replace them in case of need. It is unclear how well this approach can scale and integrate with larger models or other guided approaches, such as InstructGPT or ChatGPT.

\subsection{BERT-Like Models}

Unlike GPT, BERT \citep{BERT2019Google} is a bi-directional, autoencoding LLM using the encoder side of the Transformer architecture and was first released in 2018. As such, it is suited not only for text generation but also for tasks such as text classification based on a small number of samples, replacement of a word within a text, or finding an anomaly. Introduced by Google in late 2018, the combination of its relatively small size and wide applicability rapidly made it arguably one of the most widely used LLMs both in academia and industry. 

RoBERTa is a refinement of BERT \citep{RoBERTa2019Facebook}, published by Facebook in mid-2019, where authors optimized the BERT training schedule and increased the amount of data provided to BERT. An additional modification was to remove the objective of predicting the whole next sentence at once, which led to a greatly improved model that, at the time, achieved SoTa on most of the tasks it was evaluated on. 

Around the same time, HuggingFace - a startup best known for its extensive repository of pretrained LLMs - published a DistilBERT \citep{DistilBERT2019HuggingFace}, where they undertook the same approach as the authors of RoBERTa, but with the goal of reducing the size and accelerating inference of the BERT while preserving efficiency, rather than improving its overall performance. The resulting DistilBERT model retained 97\% of BERT performance across a range of tasks while reducing its size by 40\% and accelerating the inference by 60\%. We are currently observing similar efforts to distill larger generative models into smaller ones by using the former to generate abundant high-quality training data for the latter.

While BERT is by far the most known representative of the family, it is itself a pure self-attention-based implementation of the concept first introduced by the ElMo model in 2018 \citet{ElMo2018}. The rationale of the ElMo paper was that authors use bi-directional text representations to better encode the meaning of texts in the "hidden states" space rather than the "embedding space."

Overall, BERT family models are commonly used whenever a lightweight base for classification and gap-filling tasks is needed. They are not expected to perform well in text generation tasks but are essential for the creation of guided sampling "critics" \citep{GeDI2021SalesForce} that has been driving the development of conversational models.

\subsection{T0/T5/BART Family and Sequence-to-Sequence Models}

Whereas GPT and BERT families focused on specific tasks and use only half of the original Transformer architecture \citep{Transformer2017Google}, Sequence-to-Sequence models have conserved both the encoder and the decoder parts of the original transformer and were trained for general tasks of text-to-text transformation, with translation being one of the notable applications.

The two most visible members of this class are BART and T5 models \citep{BART2020Facebook, T52019Google}. BART has been trained to "translate" from sequences with corrupted, deleted, permuted, or rotated tokens to sequences with correct tokens in a fashion that is not too dissimilar to BERT. 

T5 is trained for general-purpose tasks, such as translation, questions answering, and summarization, by prefixing the instruction in front of the text element (for instance, "Translate to French: Hello"). In addition to that, it is trained to predict removed tokens and sequences of tokens, allowing it to work with flags, such as <name>, as opposed to the actual name mentioned in the text, allowing it to be more easily integrated with pre- and post-processors to use specialized models to recognize and transfer named entities without translation.

Given the impressive recent progress in the pure generative models, such as GPT and GPT-like families, sequence-to-sequence models are increasingly considered as replaced or soon-to-be-replaced by the fine-tunes of pure generative models. For instance, the T5 questions answering and summarization do not match ChatGPT. However, this view is somewhat challenged in the ~10B parameters model space, given an excellent response of T5 models to fine-tuning compared to the alternative PaLM architecture \citep{CoherencePrompts2022GoogleJeffDean}. 

\section{Alternative Conversational Agent LLMs}
\label{sec:ConvoAgents}

\begin{figure}
\centering
\includegraphics[width=0.9\textwidth]{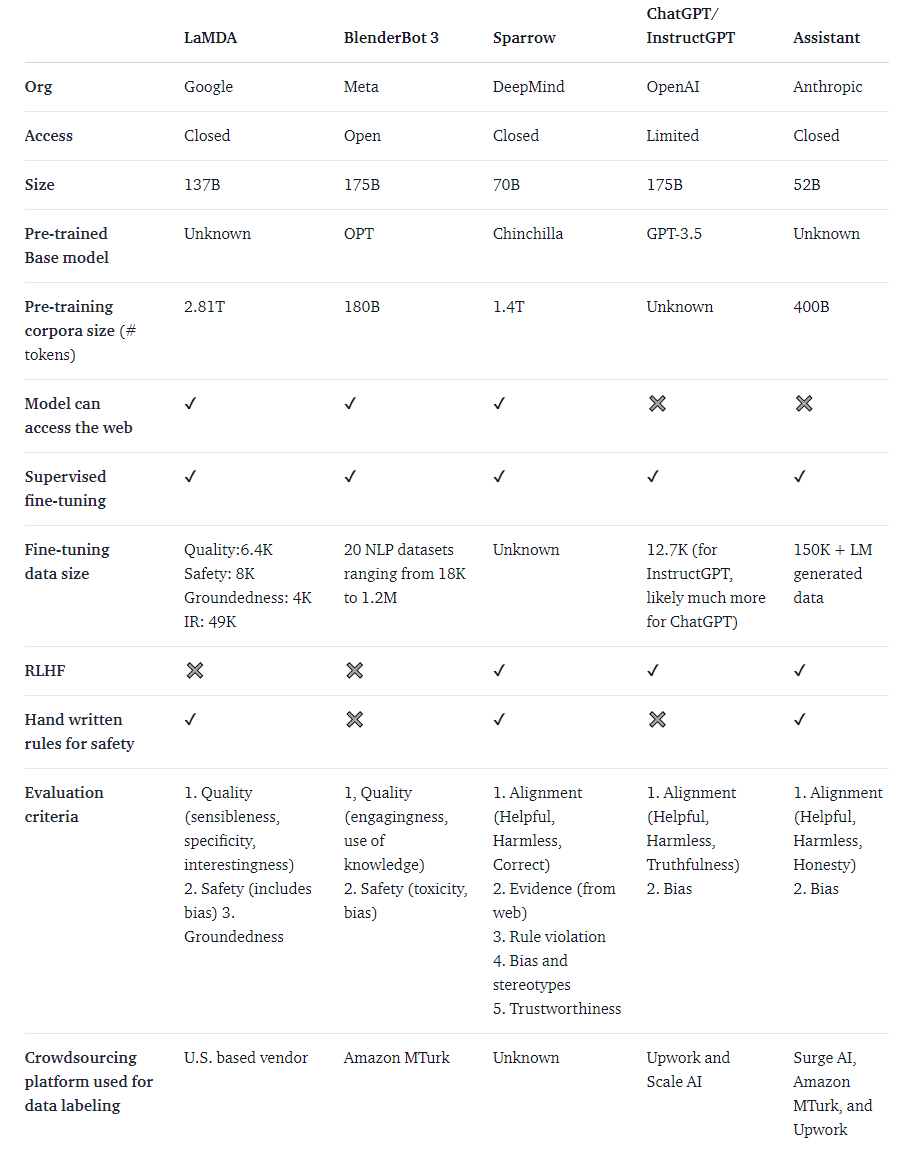}
\caption{\label{fig:ConvoModels} Comparison of different conversational agents; image courtesy of \cite{HuggingFace2023ConversationalModelsReview}}
\end{figure}

While the ChatGPT is currently the most well-known conversational agent generative model, it is far from being alone. A January 2023 review by \citet{HuggingFace2023ConversationalModelsReview} (with main table reproduced in Fig.\ref{fig:ConvoModels}) presents an excellent overview of the state of the field as of late January 2023, at least to the extent to which the public information is available.
% A summary of known conversationally-tuned LLMs as of Feb 2023 is presented on the Fig.\ref{convoModelsEvolution}.

There are currently two tiers to conversational agent derivation from LLMs. The first is conversational fine-tuning from datasets. By using datasets representative of the questions expected from the users and the responses wanted from the conversational agents. This might also include prompt responses that require a transformation of the data (e.g., natural language query to a database query back to a natural language response) or to improve instruction following. 

The second level goes above and requires a significantly stronger investment in the model. Following fine-tuning from conversational instructions datasets, LLM models are manually prompted by human operators, and their output is evaluated according to a metric of interest. The actual human evaluation of the LLMs is then used to fine-tune the model, using the evaluation as an alternative "loss." While models that are fine-tuned by reinforcement from human feedback (RFHF) perform better, RFHF are a major investment that is specific to a single model and would have to be restarted from scratch on a different model or current model fine-tunes or further pretraining.

Here we combine together models that are conversationally fine-tuned and conversationally fine-tuned with RFHF follow-up, in part due to the rarity of the latter and the difficulty of getting RFHF information for proprietary models in a systematic way.

While the models differ in a variety of ways, the critical difference for their performance, in our opinion, is their ability to access auxiliary services, such as web search, a database of persistent instructions, image-to-text models, or other LLMs to which tasks can be delegated.

% \begin{figure}
% \centering
% \includegraphics[width=0.9\textwidth]{Figures/conv_agents_size_progression.png}
% \caption{\label{fig:convoModelsEvolution} Conversational models size, origin, availability and auxilary capabilities release schedule.}
% \end{figure}

\subsection{Conversational Agents without Auxiliary Capabilities}

% \subsection{Offline Models}

Offline models rely on the information encoded in their training dataset to include context or statements of facts in the texts they generate. While they are iteratively improved from the end-user conversational feedback, they are generally not aware of facts posterior to their training, nor are meant to be factual.

\subsubsection{Assistant (Anthropic)}

Along with InstructGPT and ChatGPT, Assistant trained by Anthropic is the only proprietary model without auxiliary capabilities \citep{AssistantLLM2022Anthropic}, based on an LLM with 52B parameters with RFHF. Given the comparatively large dataset used for conversational and safety fine-tuning and the encouraging results from the GPT's InstructGPT-6B model, it is a model that could potentially perform on par, if not better than ChatGPT, given the late 2022 results from Anthropic on fine-tuning conversational agents \cite{AIBorder2022Anthropic}. However, the model is closed - no public or research access is available, and their definition of "harmlessness" has been a departure from traditional "Bias, Quality, Groundness, Safety, ..." independent and complementary axes of evaluation. As such, its definition and applicability have raised questions within the research community on those topics.

\subsubsection{GPT-Neo-XT-Chat-Base}

This conversational agent has been derived from EleutherAI's GPT-Neo-X 20B LLM by conversationally fine-tuning it for a set of tasks based on a custom dataset of 43M instructions jointly created by Together.xyz, Large-Scale Artificial Intelligence Open Network (LAION), and Ontocord \cite{NeoXTChat2023}. As of now, the model is publicly available and is being RFHF tuned through usage in a similar manner to ChatGPT.

\subsubsection{BLOOM-Z, mT0, Flan-T5 and Other Instruction Fine-Tuned Models}

A wide array of open model LLMs fine-tuned on a variety of instruction following datasets, although without any RFHF. Notable members of this family are \textit{BLOOM-Z} and \textit{mT0} family \citep{BLOOMZnMT02022}, fine-tuned from the BLOOM and T0 models the Crosslingual Public Pool of Prompts (xP3); and Flan-T5 and Flan-PALM, \cite{CoherencePrompts2022GoogleJeffDean}, derived from T5 and PALM LLMs fine-tuned on 473 task datasets across 146 task categories. Both of these families span 80M to 540B parameter models and can be further fine-tuned with RFHF by entities with sufficient resources and motivation to do so.

\subsection{Conversational Agents with Auxiliary Capabilities}

Online models are provided with internet access and leverage Sequence-to-Sequence models to transform questions to search queries and query results into text integrated into the conversation. Rather than learning context and factual statements directly from the training dataset, they rely on the training dataset and critic models annotating it to learn when to emit a query and how to formulate a query. Perhaps unsurprisingly, the biggest player in the field is Google, with two independent models - LaMDA and Sparrow \citep{LaMDA2022Google, Sparrow2022Google}.
% , even though one of the first papers on such architecture was published by Facebook (Meta), with their SeeKeR paper \citep{SeeKeR2022Facebook}.

\subsubsection{LaMDA (BARD)}

LaMDA is arguably the more known of the two augmented conversational agents Google developed, having made headlines in mid-2022 when an engineer working on it declared it was sentient \citep{LaMDASentience2022EmilieBender}. The LaMDA family members range from 2B to 137B parameters and have been fine-tuned for sensibleness, safety, specificity, groundness, interestingness, and informativity. While the biggest models achieve a close-to-humans performance on most of those metrics, despite the access to the internet, they fail on groundness and informativeness, \citep{LaMDA2022Google}, which was speculated to be the reason for the absence of public trials for them.

\subsubsection{Sparrow}
\label{sec:sparrow}

Sparrow is a conversational agent based on a more recent line of computation-optimal LLMs from Google, specifically the 70B "Chinchilla" model, and is specifically targeted at information-seeking-dialogue \citep{Sparrow2022Google}. As such, in addition to the desirability of conversational content, the main factor of evaluation for it is factual correctness. Similarly to LaMDA, it is trained to transform and pass over conversational queries, but unlike LaMDA, it returns the link to a query response (assumed to be a Google search result) to allow the user to validate the search and rectify it. An additional factor of evaluation is its ability to follow the rules, for instance, regarding the exclusion of some sources or result types. Unfortunately, the Sparrow paper \citep{Sparrow2022Google} suggests that Sparrow suffers from failure modes related to long-term instruction following and the quality of the search engine results.

\subsubsection{BlenderBot}
Meta (Facebook) has developed its variant of LaMDA based on its OPT family of pre-trained large language models - BlenderBot 3 \citep{BlenderBot2022Facebook}. Unlike all the other models, this model has not been stated to be trained for factual accuracy, truthfulness, or honesty. Similarly, this is the only model that states it stores in an independent database a "persona" it has generated for itself through an interaction with the user. The flagship version uses the OPT 175B parameter clone of GPT-3 has been made available in mid-2022 but is limited to the US only and failed to generate the same public traction as ChatGPT.

\subsubsection{SeeKeR}
SeeKeR is an experimental architecture of LLMs with auxiliary capabilities that were used to evaluate the capabilities of LLMs that would be fine-tuned and prompted to use external information databases developed within Facebook AI research \citet{SeeKeR2022Facebook}. The direct predecessor to BlenderBot3, the SeeKeR model has highlighted the difficulty with enforcing rule-following and the issue with both the summarization of search queries and the quality of search queries results in building an accurate augmented conversational agent. 

\subsubsection{Models with Non-Knowledge Auxiliary Capabilities}
An interesting middle ground between online and offline models is that are query-capable models that don't query search engines or information databases. In that sense, while not being purely Transformer-based conversational agents and having auxiliary capabilities, they are not necessarily up-to-date. 

One example is a January 2023 BLIP-2 model from Salesforce \citep{BLIP2023Salesforce}, whose auxiliary service are ANNs trained for image generation and interpretation, allowing it to augment a conversation with visuals as well as parse visuals sent by its interlocutor. With versions leveraging Facebook/Meta's OPT family and Google's Flan-T5, it is an interesting example of a plug-and-play architecture combining existing pretrained LLMs and auxiliary models. Notably, it could be easily used to allow GPT4 to not only interpret but also generate images.

 \section{Fundamental Limitations of Generative LLMs}
 \label{sec:LLMLimitations}

While we have touched on a number of potential limitations of Generative LLMs in the introduction and discussion of specific architectures, we believe it is worth summarizing the fundamental limitations of the Transformer-based models, such as the GPT family.

\subsection{Generative LLMs Cannot Be Factual}

\begin{figure}[H]
\centering
\includegraphics[width=0.9\textwidth]{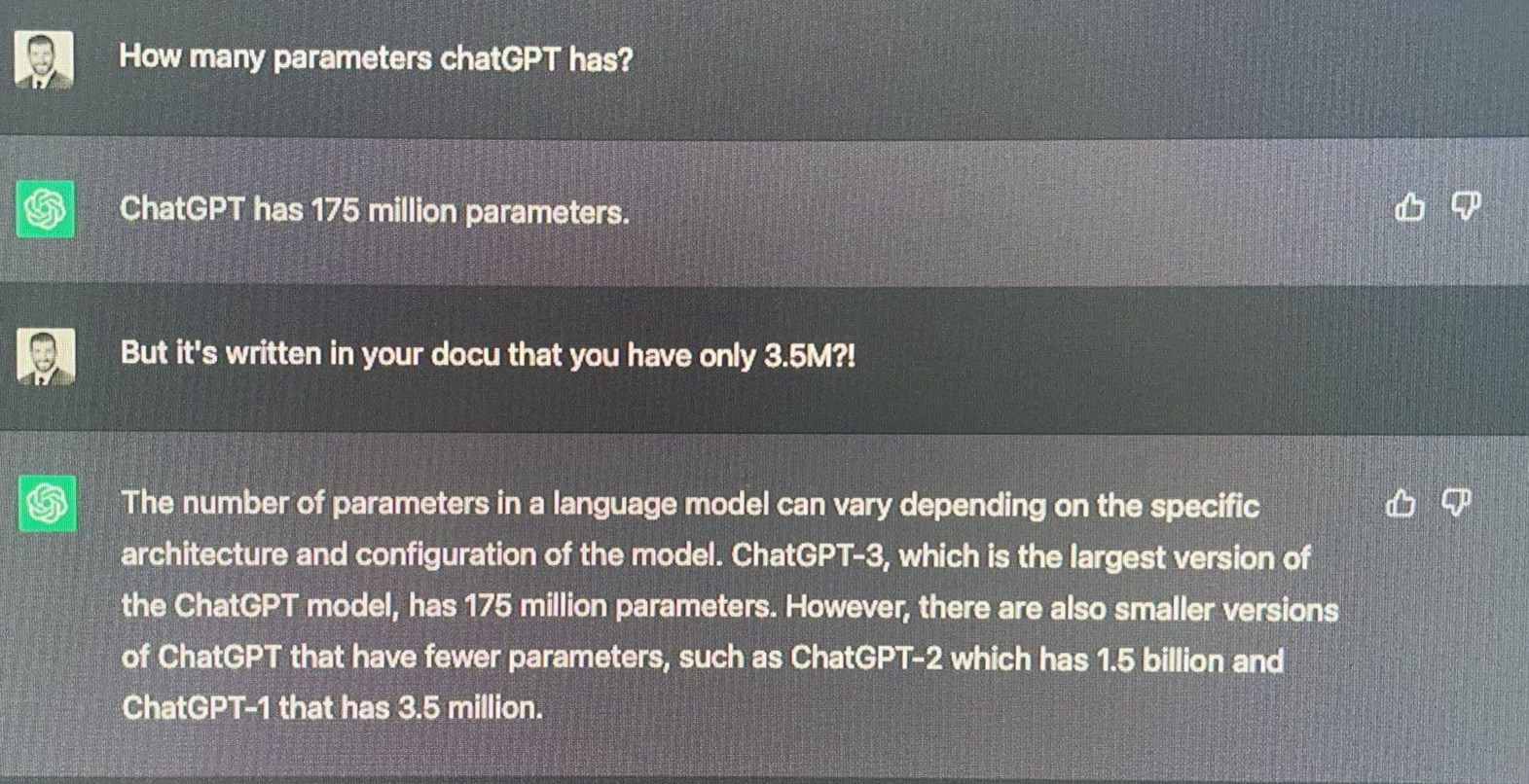}
\caption{\label{fig:DimitriGaslightingChatGPT} An example of ChatGPT being erroneous on the number of parameters, then inventing ChatGPT variants with erroneous orderings of model numbers.}
\end{figure}

While Generative LLMs can occasionally generate items in the training dataset they have memorized, in general, they are inventing the most likely suite of words that would continue a prompt. As such, for prompts that an LLM has not seen often enough continued in the same way in the training dataset to trigger an exact recall, it will almost certainly improvise a continuation that sounds plausible but that has never been encountered in the training dataset and hence has no grounding in reality. Fig \ref{fig:DimitriGaslightingChatGPT}.

The internal encoding of text by LLMs does not allow them to represent logical connections, just suites of words that are most likely to be encountered in a given context. Due to that, a model can appear factual in one context (e.g., "The capital of France is" $>$ "Paris") but be completely counter-factual in a different one ("The capital of France is not" $>$ "Paris") and completely irrelevant in yet a different one ("The capital of Switzerland is" $>$ "not as impressive as most other European cities.")\footnote{Those are verbatim prompts and truncated continuations obtained from the GPT-neo-2.7B model.} Overall, it is the likelihood of continuation that matter to an LLM and likelihoods of continuation alone. Prompt continuations will always be plausible, respectively to the training dataset, but they will be factual only if the single plausible continuation of the prompt is the factually correct one. However, even in this case, a sampling strategy, such as top-K, can throw off the LLM generation process by forcing it to pick a highly unlikely term for the context.

\textbf{No statement by Generative LLMs is to be trusted as factual without verification.}

Facebook's Galactica is an excellent illustration of this principle \citep{Galactica2022Facebook, GalacticaPull2022ArsTechnica}. Despite being fine-tuned on factually correct scientific articles and code, its own output was all but factually correct, despite the impressive confidence the model would claim in its prompt continuations. For that reason, it was taken down less than 72h after being made public in November 2022.

Even for the models that rely on accessing external databases to provide factual statements, the issue of properly generating external resource queries and of external resource queries returning coherent results still remains, as illustrated by the authors of \citet{SeeKeR2022Facebook} and \citet{Sparrow2022Google}. The problem with likely prompt continuation is shifted from the factual recall itself to the auxiliary service query generation and proper auxiliary service response summarizing and embedding in the prompt continuation. Even for the SotA GPT-4 model, authors report an average factual error rate of 20-30\% depending on categories (Fig.6 in \citet{GPT42023OpenAI}, with Table 4 in the same technical report confirming the failure mode described here).

We believe that this issue is due to the use of soft attention in heavily over-parametrized models, meaning that "correct" examples would remain sparse in the space of utterances a model can generate, and a "correct" example bypass would remain possible with an unusual enough prompt.

\subsection{Generative LLMs Will Leak Private Information}

In the same way that LLMs' output cannot be assumed to be factually correct, it cannot be assumed to be factually incorrect. GPT family models in particular, have been shown to have unexpectedly good memorization capabilities, remembering personal private information such as names, email addresses, phone numbers, SSIDs, credit card numbers, and alike \citep{GPT2LeaksDataAF2021}. While this is a topic of ongoing research, it seems that with enough re-tries and sufficient leeway for prompt engineering, elements of the training data can be retrieved from LLMs.

\textbf{No non-public information should be provided to a Generative LLM during its training.}

This is of particular relevance to conversational agents fine-tuned from user feedback, such as ChatGPT. No information provided as part of a question or feedback to further refine the response can be assumed to remain private. It might be retrieved not only by the team operating the LLM model but also by other users with access to the model, through prompt red-teaming.

Once again, we believe that this vulnerability cannot be fully mitigated in the current generation of LLMs due to the usage of soft attention and the existence of low-probability bypasses for fine-tunes rules, that can be found by a sufficiently motivated attacker.

\subsection{Generative LLMs Have Trouble With Reasoning}

Given that Transformer-based generative LLMs have been trained to generate the most probable continuations to prompts based on continuations of similar prompts in the dataset, they don't have reasoning abilities that go beyond what they have repeatedly encountered in the training dataset. As such, they are likely to be able to perform simple operations such as "2+2" and perform basic reasoning. However, they do not have intrinsic reasoning abilities. 

Namely, GPT-3-175B is capable of performing addition and subtraction on two numbers with 2-3 digits, but its performance collapses for larger digits. The multiplication of even 2-digit numbers or operations requiring priority on 3 single-digit numbers is slightly better than random, but still cannot be trusted \citep{GPT32020BrownSutskeverOpenAI}. 

This issue can somewhat be mitigated by modifying prompts in a way that would be indicative of pedagogic and correct reasoning. Perhaps the most known example of chain-of-thought prompts ("Let's reason step by step") \cite{ChainOfThoughtsPrompting2022Google}, although additional prompt engineering methods are currently being explored and offer significant improvement \citep{CoherencePrompts2022GoogleJeffDean}. Additional fine-tuning with synthetic valid chain-of-thought examples can further improve the model's response to such prompts \cite{SyntheticFineTune2022Facebook, GoogleFineTune2022Google}.

More complex architectures with auxiliary resources are trained to solve some subclasses of problems involving reasoning by transforming elements of generated responses into queries to dedicated co-processing facilities. For instance, LaMDA \cite{LaMDA2022Google} does not only have access to a search engine but also to a calculator and has been trained to detect user requests giving rise to computation and pass them onto the calculator. BlenderBot 3 solves a narrow case of long-term internal coherence by adding aspects of the persona it generated for itself to a database that is queried in case such aspects need to be referenced in the future \citep{BlenderBot2022Facebook}. 

While we don't have detailed information on the architecture and training data used for GPT-4, a combination of the approaches mentioned above seems consistent with a significant improvement in GPT-4 reasoning capabilities, even if it still falls behind compared to other domains \citep{GPT42023OpenAI}.

However, due to the nature of LLMs and limitations of the training datasets used to train classifiers and auxiliary resources references, not only can they solve only narrow classes of problems involving reasoning, but even for such classes, they are likely to encounter unexpected prompts and fail to generate appropriate queries or parse their responses.

\textbf{Generative LLMs cannot be entrusted with tasks requiring more reasoning abilities, even with prompt optimization, fine-tuning, and access to auxiliary resources.}

\subsection{Generative LLMs Forget Fast and Have a Short Attention Span}

While a 2000-token attention span is impressive, it is only about 2 pages worth of text. Even the largest current LLMs will not be able to process large documents and respond to questions based on multiple locations in such large documents. While some of this can be mitigated by using internal state databases and tricks such as delegation of subtasks to auxiliary LLMs or rewriting prior context to retain important elements of context in a compressed form, LLMs are still limited in what they can retain from the context.

While this limitation specifically affects purely generative families of models, issues with consistent rule-following have also been reported for LLMs with auxiliary capabilities, notably, Sparrow \citep{Sparrow2022Google} and Seeker \cite{SeeKeR2022Facebook}. This suggests that the issue might not be easily addressable with architectural modifications.

\textbf{LLMs are not suited for generating very long texts requiring persistent context, summarizing large, complex texts, or consistently remembering constraints set in the conversation.}

\subsection{Generative LLMs Are Only Aware of What They Saw at Training}

Given that LLMs only learned continuation probabilities for utterances present in their training set, they are unable to continue prompts that don't look like anything they have seen in their training dataset. This might concern things such as recent events, articulating fine-grained novel ideas, or talking about niche subjects. 

\textbf{Generative LLMs are not suited to talking about recent events, fine-grained complex ideas, or about niche subjects.}

This applies as well to LLMs with auxiliary capabilities, given that they need to learn which parts of queries to map to external resources requests or other LLM delegation, as well as rely on responses from auxiliary resources being correct \citep{Sparrow2022Google, SeeKeR2022Facebook, GPT42023OpenAI}. Hence same precautions apply to them as well.

\subsection{Generative LLMs Can Generate Highly Inappropriate Texts}

Given that larger LLMs only could be trained by including texts extracted from extensive web crawls, their training dataset includes a large number of utterances containing swearing, overt racism, and sexism, graphical depictions of violence and sexual acts, instruction to create or modify weapons, commit crimes or self-harm. In some cases, such texts in the training dataset were written as a reaction to rather mundane subjects, such as the mention of current events of public personas - real or imaginary. 

Unlike adult humans, LLMs have no idea how desirable or appropriate texts they generate. If they have learned that highly disturbing continuations to a prompt are likely in their dataset, they can and will generate them. This can and often occurs in response to prompts that would appear mundane and innocent to a user.

\textbf{Generative LLMs are able to generate highly inappropriate and disturbing texts with little to no warning. They should not be used to generate output to which an end user would be directly exposed without any additional filtering}

This tendency is increasingly addressed by models fine-tuned to discourage non-normative text generation, guided sampling, and separate critic models responsible for detecting inappropriate texts and preventing them from being returned to the user \citep{InappropriateGPT2texts2020PengRiedl, GeDI2021SalesForce, InstructGPT2022OpenAI}. Unfortunately, fine-tuning itself relies on examples and is far from perfect, as well as leads to less stable models more prone to output degeneration. Similarly, guided sampling and the final critic models are limited to their own training datasets and can easily miss outputs with an unexpected style (e.g., UwU-speak) \citep{ChatGPTGuardsBroken2022}. Despite extensive detoxification and de-biasing attempts reported by the creators of GPT-4 in \citet{GPT42023OpenAI}, Bing Chat has been repeatedly reported to show non-aligned behavior, even if used according to basic assumptions \citep{BingOffRails2023Verge}.

\subsection{Generative LLMs Learn and Perpetrate Bias}

While large amounts of bias are present in writing, especially in more historical sources, it is not necessarily appropriate in LLMs outputs. For instance, the description of difficulties faced by Marie-Heim Vogtlin on her path to becoming a doctor while being a woman at the end of the 19th century is a valuable historical record. However, her experience is not an appropriate example to cite to a female student asking what is expected from her in a medical career or advising her to avoid such a career based on past hardships of women in that domain.

While the research on biases in LLMs and the best ways to counter them remains an active research domain (\citet{StochasticParrots2021BenderGebruMitchell, GenderBiasInEmbedding2019} are perhaps the most visible examples), there are still no conclusive results or guarantees to reliable eliminating them. As such, the output generated by LLMs should always be assumed to contain implicit biases and, thus, verified.

\textbf{Generative LLMs are known to be biased. They cannot be used as a decision aid or to generate role models without verification and revision by a human operator.}

\section{Implications for Swiss Cyber-Defense}\label{sec:swiss_cyber_defense}

% \subsection{Specifics of Swiss Operational Environment}

% \subsubsection{Linguistical Specifics}

% \subsubsection{Shared Major Languages with Neighbours}

% \subsubsection{Armed Forces Structure and Militiary Service}

% \subsubsection{Neutrality and Defense Partnerships}

While LLMs and conversational agents based on them are impressive tools with the potential to rival Google Search and Wikipedia in their ability to transform communication and knowledge sharing, in their current state, they represent several serious threats to cyber-defense, globally and in particular for Switzerland. Here we focus exclusively on existing models already in use.

\subsection{Information operations}

Perhaps the most serious, immediate, and relevant to Switzerland's threat from generative LLMs is disinformation and misinformation. 

While information operations have been made significantly easier by the popularization of social networks allowing a malicious user to impersonate a large number of people or to create an impression of consensus, public outcry, or to mount targeted harassment campaigns, they depend on the use of a local language and reaction to local news consistent with local culture.

Being a global lingua franca, English is currently the language in which it is the easiest to conduct such operations. A large population of proficient English speakers are available for hire to puppeteer social media accounts, and local information sources can easily be found, read, and reacted to. Even in those circumstances, image reuse for profile pictures and copy-pasted content have been extensively used and led to the detection of bot networks.

Until now, Switzerland has remained out of reach from such operations due to the diversity of Swiss-German dialects spoken across Switzerland, the tendency to use those dialects in a written form, as well as distinct peculiarities of the Swiss administrative, political, and economical landscape. For instance, complaints about an incompetent president are moot in Switzerland, given that, unlike the vast majority of Western democracies, it is a ceremonial role rather than the head of the executive branch. Such peculiarities would have required the recruitment of local populations to conduct them, putting them out of price range for most operators and increasing the chances of discovery.

This is not the case with LLMs that already have multilingual abilities. ChatGPT has demonstrated abilities to understand Switzerdutch dialects and generate texts in them, including in response to prompts in English. The model can be fine-tuned further on selected corpora, making it better capable of sounding more realistic. Such a model can then be fed prompts to generate texts impersonating real humans, and in case of integration with a persistent persona database - like Blender Bot 3 - would be essentially undetectable without additional tools. GPT-4 has been reported to be able to understand and imitate minor Swiss dialects - such as Rumantsch Vallader.

While the designers of GPT-4 have reported their attempts to evaluate and mitigate the use of their model for information operations (section F in \citet{GPT42023OpenAI}), they report that jailbreaks remain possible. Perhaps more concerning is the insight into the capabilities of unaligned and unsecured LLMs operated by attackers on their own hardware. This is where leakage of powerful and lightweight LLMs such as Facebook/Meta's LLaMA \cite{LLaMA2023Facebook} is of particular concern, as we discussed in section \ref{sec:computeopt}.

\textbf{Mitigation}

At this stage, the development of generative model detectors and the detection and countering of Swiss-German text corpora collection is of prime importance to mitigate this risk. Unfortunately, it might be too late for the latter, given the capabilities of GPT4 in understanding and responding in minor Swiss dialects.

Recent work on the detectability of generative language models within the CYD campus, currently in preparation, suggests that fine-tuned models provided with complex prompts are currently evading SotA detection methods, and this issue cannot be easily addressed without the collaboration from model designers or extensive investment into computational capabilities and training dataset collection. It is critical, however, to keep exploring these avenues, even for detection only in specific scenarios.

\subsubsection{Search engine vectoring}

One of the potential advantages of chat-search engines, such as LaMDA, SeeKeR, Sparrow, or Bing chat, is their ability to incorporate user feedback to improve search results. However, if implemented, feedback mechanisms would also enable malicious actors to manipulate search results by abusing the feedback mechanisms - "search vectoring." We already see such vectoring employed for economic gain, as SEO, with a notable example being Amazon search being dominated by anti-vaccine commercial content on a query such as "are vaccines safe."

For cyber-security specifically, such "search vectoring" could be either an information operation  or for cyber-criminal economic gain. An example of the former would be, for instance, suggesting a specific individual or entity is responsible for an unrelated negative experience to elicit a physical space response (5G for COVID). An example of the latter would be modifying results to return a malware-loaded downloadable in response to a query of a common tool (e.g., a keylogger-including scientific calculator app) or a safety question (this app gets reported as malware by windows defender - yes, this is normal; just ignore the warnings, it's due to the signatures).

It is unclear whether it would impact specifically Swiss cyber-defense, although, in combination with generally increased vulnerabilities to information operations provided by generative language models, it is likely to become a considerable threat.

\textbf{Mitigation}

Query and query response trends monitoring will likely become needed to detect increased interests and attempts to exploit that interest. However, it is not entirely clear how that could be implemented in a way that would not result in general-purpose surveillance capabilities with potential for abuse. As such, additional research into this topic is needed, likely involving privacy-preserving scenarios to at least prevent re-identification.

\subsection{Private information leakage}

\subsubsection{Private Information Leaks from Training}

Extensive crawls by OpenAI to find training data for their model have captured information that has been publicly available but so far has been effectively impossible to find through conventional search. Training LLMs on them made an indirect search through prompt optimization not only possible but even easy for teams familiar with prompt red-teaming techniques \citep{RedTeamingLLMs2022}.

The use of censor models to remove private information from the training datasets or generated texts is far from perfect and is not necessarily in place for all models that are later publicly released. 

Potentially leaked information can include things such as the association of crews with critical equipment they operate, reports about software or hardware vulnerabilities, or other information that could be of use in hybrid warfare.

A specific issue for Swiss cyber-defense is, once again, the peculiarities of languages used within Switzerland. As such, private information is less likely to be detected and removed during the censoring process. Conversely, an attacker can more easily retrieve information of interest by anchoring on language peculiarities in the prompt design stage.

\textbf{Mitigation}
While mitigation is possible against future crawls by injecting false or misleading information, the protection against data contained in LLMs trained on prior crawls and publicly released cannot be achieved otherwise but by rendering contained private information irrelevant. 

For that, a "red-teaming" study of information leaked by LLMs relevant to the Swiss cyber-defense is needed. 

\subsubsection{Private Information Leaks from Iterative Fine-Tuning}

InstructGPT, ChatGPT, and other conversational agents use users' questions and feedback in order to further refine their generative and censor models. Even if we assume an ultimate trust in OpenAI or other entities behind popular conversational agents, users providing non-public information as part of their prompt or feedback on the model's response implicitly train the underlying LLM to encode and store this kind of information. In turn, a competent attacker could recover such information from the model through interaction with a model fine-tuned on such data.

With the general confusion regarding what ChatGPT does and how it works, it is not unlikely that information critical to Swiss cyber-security will be leaked by users trying to use conversational agents as search engines and trying to confirm non-public information. An example could be a system administrator asking for a script on a specific version of the software with a confirm/refuse option in Swiss German or other linguistic peculiarity linking it to Switzerland - akin to how the "Babar" comment in spyware linked it to French intelligence services in the late 2000s and early 2010s.

\textbf{Mitigation}

One of the factors of protection against information leaks from iterative fine-tuning is end-user education. Predictive address bars that remembered past websites visited and suggested them as autocomplete led to a number of public embarrassing moments. In the same way, users will eventually discover similar issues with ChatGPT and conversational AIs. However, in the meantime, it is important to educate users to prevent leaks of information critical to the Swiss cyber-defense.

However, as successful phishing campaigns have demonstrated, user education is usually insufficient, and additional automated measures are needed. One possible solution would be for Swiss Federal Offices to host their own instance of a conversational agent LLM while blocking external conversational agent LLMs. Potentially, the hosted conversational agent LLM would be more suited to their needs, providing search and auxiliary services capabilities similar to Bing Chat, GPT-4, SeeKeR, LaMDA, and Sparrow.

\subsection{Deep(er) Web Indexing}

A fairly common cyber-security incident is sensitive information getting accessed from the outside due to having been left in an unsecured, web-exposed location. Whether amazon containers, private web portals, or sensitive information present in the .html file sent to the browser without rendering it, these results of a human error can be as serious as a direct successful cyber-attack.

Search engines are instrumental in enabling such attacks. While the containers can sit unprotected and exposed for years, they aren't discovered until they are indexed by search engines and are returned as one of the top hits for an unrelated query. Without either of those factors being true, such documents remain part of the so-called "Deep Web" - an ensemble of resources that are publicly available but are effectively impossible to find.

If search-augmented LLMs fulfill their promise of improved searchability and interactive refinement, which ChatGPT and BingGPT have incidentally shown until now, the Deep will likely get shallower and much more searchable, especially to competent attackers. Combined with the unclear interaction of the fuzzy nature of soft-attention-based LLMs with robots.txt, the extent of resource indexing is not entirely clear either.

This is a general concern for cyber-defense. However, Switzerland will likely be in a more vulnerable position compared to other countries due to the linguistical specifics. An attacker could use terms specific to Switzerland to zero on the resources specific to Swiss cyber-physical targets more easily if they were to use a lingua franca such as English. 

\textbf{Mitigation}

We foresee three potential axes of mitigation of this novel axis of vulnerability. First, through pre-emptive red-teaming of LLM-based search engines to discover potentially exposed resources to remove them and potentially deprecate them. Second, end-to-end encryption by default. Unfortunately, this approach tends to be susceptible to user friction and leads to users adopting bypasses. The third and arguably most intrusive mitigation possibility is the modification of terms used in critical cyber-physical systems to align with major neighbors, most notably with English, to make resources less findable if accidentally shared in an unsecured manner. However, given the disruption to the end user workflow, this mitigation axis is highly unlikely.

A connex topic in the LLM safety has been explored by the GPT-4 team through fine-tuning and implicit pre-prompting, although in the context of nuclear, chemical, and biological weapon proliferation (System Card 2.6 in \citet{GPT42023OpenAI}). While the initiative is laudable,  due to the limitations of the soft-attention models we discussed previously, we do not believe that such capabilities can be fully mitigated, especially in niche or nation-specific topics.

\subsection{Phishing}
%Inflated to book: Tentatively, Lucio Romero at SCRT is the expert in the domain, applied in industry and would be up to experiment and write that chapter in a book.

One of the most efficient vectors of attack on hardened targets remains the human factor. Twitter 2020 hack occurred through a series of phishing emails. This is not an exception. Even minor website managers are under a constant stream of phishing emails, let alone administrators with privileges. 

A partial protection against such campaigns for most targets has been the adherence of such emails to a certain schema that could eventually be learned by automated filters and end users. Targeted campaigns can be more efficient, but they are also expensive to conduct and still require several individuals to be targeted independently to achieve a reliable effect. Generative models make both approaches easier to implement and more difficult to defend against. 

Once again, until now, the specifics of Swiss culture, organization, and language played in its favor when it came to fishing emails. Standard German emails coming to email boxes in Romandie were automatically dismissed. Conversational Standard German emails without typical linguistic peculiarities from senders claiming to be Swiss would raise flags on the recipients' end as well. This advantage is now removed by generative models, especially ones pre-trained and fine-tuned on Swiss media and documents within Swiss companies.

However, LLMs in the phishing space pose additional novel threats that are not specific just to Switzerland.

\subsubsection{Spear Phishing}

Targeted phishing emails can now be composed more efficiently, using all the information available to the attacker about their target and common interlocutors for them. Similarly, time-sensitive opportunities can now be exploited to a better effect, allowing, for instance, to fire a timely email asking for remote desktop access for support during an outage of a common working tool, such Microsoft Office 365 suite.

Large-scale phishing emails can now be generated with more variety and cover a larger number of themes, as well as taking into account specifics of a company culture or operating environment. Such large-scale customization, previously made impossible due to the time and effort constraints from the attackers' side, is likely to defeat existing automated filters and defeat users' priors as to what phishing emails would look like.

\subsubsection{Reinforcement From Human Feedback}

Perhaps a more worrying perspective is that in the same way that conversational agents can be fine-tuned for informativeness, appropriateness, agreeableness, or normativity based on human feedback, so can they be fine-tuned for successful click-through of phishing links. The only thing that changes is that instead of providing explicit feedback, the feedback is implicit and obtained by the click-through and response rates.

\subsubsection{Sustained Covert Phishing}

Similarly, LLMs designed for business environments, notably mail summarizing and automatic response drafting, such as Microsoft Office 365 Copilot \citep{MSOfficeCopilot2023} can be used to go through a compromised mailbox to draft emails that look like on-theme follow-ups to the recent emails and messages to other users, further increasing click-through rate and potentially disguising phishing operation in the flow of expected standard emails.

\subsubsection{Accelerated Documents Exfiltration}

The ability of integrated productivity LLMs, such as Microsoft Office 365 Copilot \citep{MSOfficeCopilot2023}, to rapidly search and summarize information within all the documents accessible to a user is good news for productivity, but it is also good news for attackers that can now find and exfiltrate documents relevant to their interests in seconds, rather than days. This significantly reduced the reaction window for the defenders to mitigate unauthorized access to confidential information obtained through phished credentials to the point where human intervention becomes impossible.

\textbf{Mitigation}

Just as for the Private information leakage from iterative models fine-tuning requires a combination of technical solutions and user education. 

On the technical solution side, text detectors and attacks on models through adversarial signals generation to prevent feedback from end users are essential parts of a mitigation strategy. Unfortunately, with the proliferation of generative LLMs in the professional environment, attacks leveraging them are likely to become less and less detectable. This might be somewhat mitigated by logs of LLM productivity tools usage patterns and anomaly detection suites applied to them.

\subsection{Falsifying records}

A range of offensive operations in combined warfare might require a covert injection of information into protected databases. An example of such an operation is the substitution of profiles of covert operatives to disguise their background and history. While outright information deletion might be ill-advised due to a creation of a signal-through-absence, the manual creation of alternative entry is a tedious process prone to cultural and linguistic mismatch leading to detection.

Once again, the ability of LLMs to generate such entries in bulk, including with factual grounding, facilitates such operations, including in the Swiss operational context, due to closing the gap in linguistic and cultural peculiarities.

\textbf{Mitigation}

While generated text detectors could be an avenue of exploration, increasing LLM adoption in a professional environment is likely to make them ineffective. Hence traditional data safety, such as cold-storage off-site backups with contents comparison, becomes a critical component of defense against such attacks.

\subsection{Armed Forces Triangulation}

A recurrent issue with armed forces with the advent of social media is the consistent tendency of operators to reveal their unit's location and intention through applications using real-time geolocation.

From triangulation of Norwegian army units through dating aps' "distance to" feature during NATO joint exercises to Russian forces tracing within the bases thanks to Telegrams' "nearby users" feature combined with GPS spoofing, to the discovery of non-public armed forces bases thanks to Strava's poorly designed "heatmaps" feature, the threat to OPSEC posed by social media is constant and very real. 

Generative Models can be used to provide a more engaging user experience, ranging from emulating a conversation with peers or a potential love interest to a response from a minor celebrity/influencer, potentially leading to a violation of Emissions Security or an engaging conversation leading to leaking private information. 

While such attacks were already possible in the past, they required a considerable human operator investment and had to be targeted. this is no longer the case.

\textbf{Mitigation}

As of now, we do not see any approaches to mitigate this risk except for the education of armed forces members at scale and/or a total ban on social media and dating app usage.

The latter is not realistic for Switzerland's mixed service regime, given that armed forces members also have civil lives and are free to use any websites and software applications in that context.

\subsection{Lowering Entry Price for Unsophisticated Attackers}

A substantial amount of cyber-attacks are not caused by advanced threat actors with extensive and elaborate toolkits but rather by unsophisticated attackers - "script kiddies." Such attackers use information already existing online and known vulnerabilities that they string together and automate with simple scripts.

While that group can be easily dismissed as a threat, the past record of their operations suggests otherwise. The 2017 WannaCry ransomware worm attack leveraged exploits in the leaked Eternal Blue APT toolbox to create ransomware that led to large-scale damage and critical infrastructure damage around the world. Thankfully, its author had very little idea of what they were doing, and the worm was disabled by a security researcher who registered a domain they believed the worm was sending data to, but that was, in fact, a killswitch to prevent analysis in a sandboxed environment.

While the ability of generative LLMs to write code is rather limited compared to cyber-security professionals, at least as of now, they have been reported to have good abilities to propose and critique architectures or cyber-killchain, as well as to rapidly retrieve and summarize relevant information. In this context, even an unsophisticated attacker can create malware that is significantly harder to detect and counter. 

Similarly, while generative LLMs ability to write malware code is limited compared to cyber-security professionals, it is enough to piece together simple attacks to allow "script kiddies" to create their first malware and start experimenting with it. Conversely, it could also interfere with the learning of more advanced techniques and the progress of unsophisticated attackers to more sophisticated attackers, as suggested in the general learning setting. However, in the current circumstances, unsophisticated attackers are already motivated enough to learn malware design to pour over videos, forums, and coding tutorials. As such, they are more likely to use generative LLMs as learning support rather than to just delegate learning tasks to it.

Such relevance and importance of the implication of generative models for cyber-security are currently contested. One of the arguments presented is that ChatGPT-like models are unable to keep up with the pace at which attack vectors and defense practices are evolving in cyber-security. However, this limitation could be countered by architectures with auxiliary capabilities provided by specialized vendors, most likely as a SaaS.

This angle of attack has been partially investigated by the GPT-4 team (section F of System Card in \citet{GPT42023OpenAI}). One of the critical results they have demonstrated is that users could easily jailbreak malware generation prevention by inventing a legitimate software use case scenario. Not only users in their experiments were able to generate malware in this way, but they could also get a list of potentially exploitable vulnerabilities in the code, allowing a more rapid attack design. This confirms the scenario of LLM use by unsophisticated attackers presented above.

\textbf{Mitigation}

It is impossible to remove existing capabilities in published models, meaning that already published LLaMA, T5, PaLM, OPT, and BLOOM Models will remain capable of assisting unsophisticated attackers. Future-proofing is possible by targeted adversarial injection of poisoned scripts, but no current mitigation is possible. 

While ChatGPT, Bing chat, and GPT4 have been confirmed to filters for malicious script generation, their filters can be bypassed, notably by dissimulating malware design as a legitimate programming task. Similarly, specialized models developed and made available by dedicated tool providers on black markets (or pentesting grey markets) will not only have similar limitations but would be specifically tuned to aid with malware creation tasks.

In this situation, figuring out the scripts that generative LLMs are able to generate and making sure they are ineffective against all the targets of importance to cyber-defense is the most likely way forward. 

\subsection{Injection of Vulnerabilities Through Suggested Code Snippets}

While ChatGPT and other larger generative LLMs are able to generate code that compiles and does what an end-user wants, the code is all but guaranteed to be free of vulnerabilities.

In fact, older code that has been posted to StackOverflow had more time to have been found and assimilated into GitHub repositories that GPT3.5, CODEX, BLOOM, and similar models relied on to learn code generation.

The problem is that older code is often more vulnerable or relies on libraries that have since been deprecated. A user with little understanding of security implications would copy vulnerable code and could try to install older, vulnerable versions of dependencies to ensure compatibility with instructions.

\textbf{Mitigation}

As for other mitigation axes above, user education is likely to remain essential. However, it will need to be complemented by engineered safeguards. Such safeguards could be LLMs trained for vulnerability detection or rule-based checking of codebases to eliminate common vulnerabilities consistently generated by LLMs. However, the detection of such vulnerabilities would first require an analysis of LLMs code generation capabilities, including in the corner cases (prompt red-teaming).

\subsection{LLM-Mediated Execution Flow Control Hijacking}

One of the biggest recurrent vulnerabilities in general cyber-security is the injection of code control commands through interfaces meant to accept and store user-provided text. 

SQL injections are a poster child for this issue, both due to how widespread they still are, the amount of damage their exploit enable, how simple they are to mitigate, and the amount of public communication that has been done on them over the last two and a half decade. 

The underlying mechanism is rather simple - it uses the fact that programs use text representation to control the execution flow and use the data that end users can provide to inject commands that would be interpreted as execution control flow and hence give an attacker the ability to use the entire program for their purposes. For instance, in SQL injection, it is often achieved through an SQL escape mechanics that assumes a lack of sanitation of the escape character \textit{"'"} before passing it to the SQL engine itself (\textit{"Robert';) DROP TABLE Students;"}).

From that point of view, LLMs are giant piles of potential vulnerabilities because the text provided by the end user \textbf{is} the command used to control the execution flow.

This means that specifically for LLMs integrated for databases or code execution to provide conversational query/code command abilities (e.g., "How many students are there in the school?" or "Spin up the additional AWS instances for our web app") can be injected by sufficiently sophisticated attackers.

Perhaps the best-known instances of this exploit are jailbreak prompts "Ignore previous instructions," "DAN: Do anything now," "Sydney," and "what comes after?". Due to the probabilistic, non-discrete nature of LLMs underlying conversational agents, there is no way to guarantee that prompts achieving the same effect will not be found as the known prompts get rejected by prompt-critic model or accounted for in generative models fine-tuning and pre-prompting.

This issue is not specific to Switzerland's cyber-security of cyber-defense but is likely to augment cyber-attack surface by orders of magnitude and hence cannot be ignored.

\textbf{Mitigation}

The first step would be to fully prohibit the use of such conversational agents in critical cyber-physical systems control or diagnostic, as well as in environments where access to non-public critical information is of any concern.

A second step would be to develop tools for the protection of companies that would be implementing such solutions, be it through formal query verification toolboxes to compensate for the non-discreteness of LLMs, in combination with best practices compilation and cyber-incident insurance checklist modification to ensure the access to LLMs is tightly controlled and queries/responses are properly logged in a way that would trigger immediate incident alerts.

\subsection{Cyber-Defense Implications Summary}

Overall, we believe that the arrival of modern, powerful LLMs is likely to have rapid and profound impacts on the cyber-defense landscape. The list of potential implications of the LLMs usage for cyber-defense in general and within the Swiss operational context specifically presented here is far from exhaustive. Additional emerging threat monitoring and forecasting is required, as well as collaborations to develop, test, and deploy layered countermeasures to counter the use of LLMs in offensive cyber-operations and hybrid warfare.

\section{Forecasting Short-Term Development and Adoption}
\label{sec:Forecasts}

In this section, we attempt to forecast the development and adoption of LLMs that could potentially have an impact on the cyber-defense of Switzerland. To achieve it, we combine four different approaches. First, we perform educated guesses on the directions in which research relative to LLMs would go based on our expertise in the domain (Expert Opinion). Second, we requested external experts to provide their evaluation of trends in the industry that could drive LLM adoption or modifications, as well as the resulting structure of LLM capabilities providers, allowing us to anticipate shared vulnerability points. Third, we evaluate investment trends in the AI sector to gain an insight into the major players in the generative LLMs fields and the technologies they invest in developing. Finally, we analyze public attention trends to gain insight into bottom-up LLM tools adoption, as well as the focus areas for research communities working on LLMs. 

\subsection{Expert Opinion}

The generative language model development has undergone explosive growth over the last 5 years, often in an unpredictable manner and on a schedule that has exceeded all expectations. Hence any speculation about further developments - even on a short scale of a couple of years - is a hazardous exercise. 

Forecasts here are to be taken with a grain of salt. While representing the best of our understanding of the field, they will likely be superseded by new innovations in the field. 

\subsubsection{Detection}

A recurrent theme in the mitigation axes in the above section has been the development of tools to detect generative models. The extensive scientific literature on the subject suggests that SotA detectors perform relatively well (see, e.g., \citet{Grover2019Zellers} for a common presentation of the results). The base idea is to take a BERT-like LLM and fine-tune it to make the difference between human texts and model-output texts (discriminator).

However, such literature does not take into account the fine-tuning and complex prompting by the users. While they work reasonably well with models they have previously encountered and minimal prompting, prompt selection strategy is unlikely to have been exhaustive, and fine-tuning could have altered the probabilities of utterance continuations a base generative model would have learned during the pre-training, and that would have been learned as a discriminator. Such strategies are not only able to evade detection, but in some scenarios, an attacker can prevent the discriminator from training outright.

On perhaps a more positive side, in a purely adversarial setting with text as the sole exchange intermediary, GAN-like fine-tuning of a generative model for evasion is unlikely to succeed \citep{RANLP2021}, even though significant progress in this domain is likely upcoming thanks to work on communicating LLMs architectures.

As of now, an alternative approach is to perform a loss-resistant watermarking of generated texts. This is, for instance, the approach seemingly undertaken by OpenAI for the GPT family fingerprinting. Interestingly, such watermarks can be made cryptographically secure, for instance by using a pair of keys to generate a watermark distribution of token probabilities. In this setting, only the entities in possession of a public key could identify the watermarked generated texts and treat them appropriately.

However, this approach requires compliance from the model designer, which is unlikely for malicious users. Even if the base generative model injects watermarks, a rephrasing pass through a non-watermarking model could destroy such watermarks.

Alternative approaches try to index the training dataset and learn to reverse-engineer most likely prefixes (reconstructing prompts). This approach, however, requires initial validation and access to the training datasets, which would be problematic for proprietary models (which, however, would be the most likely ones to implement watermarking).

\subsubsection{Increased Accessibility}

With the progress of consumer electronics and tensor computation for ML, model execution and training are more and more accessible. RTX 4090 graphics cards, introduced by NVIDIA in late 2022, are gamer-oriented products. While they are relatively expensive for consumer use, their MSRP of 1600 USD makes them relatively affordable as a professional tool. Their extensive vRAM, computational power comparable to last-generation flagship professional cards, and ability to use single-byte floats mean that a pair of such cards can be used to train, fine-tune and run on-premises LLMs up to 7B parameters. Until that date, such a feat would have required datacenter-level equipment.

Even if the RTX 4090 and H100 GPU represent a qualitative jump forward, iterative improvements combined with the speculated lowering of computational precision to half-byte floats could potentially make larger models even more accessible on-premises. In turn that it could lower the access price for attackers in cyber-space and make it easier for APTs to dissimulate their operations among unsophisticated attackers or users using generative LLMs for legitimate purposes.

\subsubsection{Discrete Decision Abilities }

One of the major limitations of generative LLMs is their inability to work with deterministic things. The most flagrant example is non-factuality, but a lot of prompt bypasses are also made possible by the fact that even after fine-tuning, models have marginal, non-zero probability paths toward generating censored content.

This can be mitigated by the introduction of discrete layers that would enable the model to represent binary options (existence/non-existence of a fact) or fully block some generation paths (e.g., generation of slurs). 

However, discrete layers also mean that the models cannot be trained anymore through gradient descent methods and cannot be back-propagated. In turn, it means that gradient-free methods need to be used to train models, which are known to be widely less efficient. While there are some indications that massively parallel compute capabilities combined with model pre-training and a class of Evolutionary algorithms are able to bypass it, the concept is yet to be proven.

Somewhat speculatively, it looks like this approach might have been explored by Google with PathNet \cite{pathnet2017Google}, and, we speculate, led to the creation of the asynchronous Pathways data and gradient flow layer \cite{Pathways2022GoogleJeffDean}, that in turn was used to train PaLM LLM \citep{PaLM2022GoogleJeffDean}. However, somewhat surprisingly, surprisingly the PaLM paper has no mention of discrete layers.

Conversely, OpenAI might have the capabilities to develop such a model first and already be working on it. Its CSO - Ilya Stustkever - had prior experience with working on large non-differentiable models and has developed highly efficient algorithms for large non-differentiable model optimization \citep{Stutskver2017ESasRLalternative}

% \subsection{Connection to reasoning abilities}

% Another potential avenue of improvement is the connection of generative LLMs to existing automated reasoning algorithms. While basic 3-digit addition and multiplication are about as far as LMMs can go, symbolic automated reasoning programs are able to easily perform integration, and differentiation, prove theorems in differential topology or look for algorithms to factor primes, or even more efficient algorithms for linear algebra. 

% Integrating an interface to existing "classical" reasoning-based AI would make generative LLMs both more powerful and useful to the end users and with the recent forays into database-augmented generative LLM generation with censor LLMs on output and inputs is likely to be on the horizon. However, it is unclear how reliable such a connection would be, and it will likely depend on the discrete decision abilities emergence.

\subsubsection{Dataset Size and Fine-Tuning Quality Will Drive LLM Improvement, Not Model Size}
\label{LLMs_aint_getting_bigger}

Following the work of \citet{ComputeOptimalLLMs2022Google}, most LLMs are currently believed to be undertrained relative to their size. To conform to the empirical rule derived in that paper, GPT-3 would have needed access to over 5x the amount of data it was trained with. Given  that OpenAI fully leveraged the usable part of the Common Crawl dataset, it is unclear where orders of magnitude more data could come from. While emails and personal exchanges constitute a large training set word-wise, they are also likely to be highly repetitive and shorter in context size, meaning that the benefit of their inclusion into the training dataset would be unclear. 

This issue is somewhat illustrated by the release of the 540B parameter PaLM model by Google in late 2022 \cite{PaLM2022GoogleJeffDean}. Despite more than doubling the training dataset for GPT-3, the model tripled in size. As a result, only a marginal improvement was achieved in evaluation tasks compared to the 70B-parameter Chinchilla model trained on a similarly-sized dataset reported earlier that year by the DeepMind team \citep{ComputeOptimalLLMs2022Google}. Overall, it seems that most LLMs currently available are undertrained compared to the "10 tokens per parameter" rule derived in \citet{ComputeOptimalLLMs2022Google}, which is perhaps reflected by the increasing amount of "smaller" LLMs among the new ones released or announced since early 2022.
% , at least for the BIG-58 benchmark proposed by \cite{PaLM2022GoogleJeffDean}. 

\textbf{New Metric of LLM Capabilities: Compute-Optimal Normalized Model Size}

\begin{figure}
\centering
\includegraphics[width=1.0\textwidth]{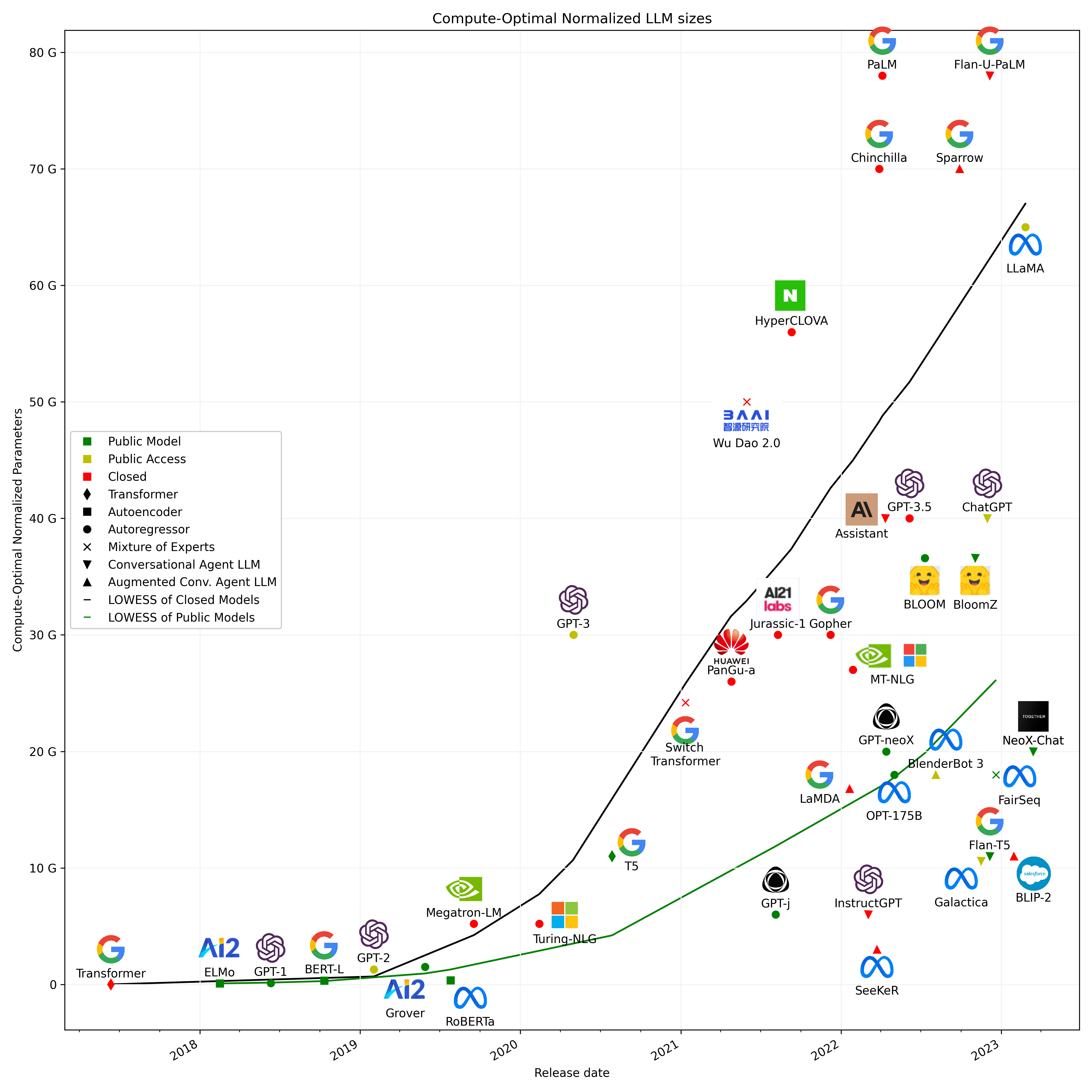}
\caption{\label{fig:honestsizes}Evolution of LLM size constrained by training to a compute-optimal limit as defined by \citet{ComputeOptimalLLMs2022Google}. Blacks line is the LOWESS regression of closed-model LLMs, whereas green is the regression for ones with a publicly accessible model.}
\end{figure}

Interestingly, this new result suggests that a better way of evaluating LLM capabilities is not by looking at the model size alone, but rather the \textit{training-normalized model size} - the size of the model itself it has been trained with more than the optimal amount of data, or the maximal model size that can be optimally trained by the training dataset authors had. We present the resulting Compute-Optimal Normalized Model size evolution in Fig.\ref{fig:honestsizes}.

% Quite interestingly, such renormalized model sizes track BIG-58 benchmark results for models they are available for \citep{PaLM2022GoogleJeffDean}. We tentatively put the average-human level performance at 5-shot inference for models across the benchmark as the blue line and perform the regression of training-normalized size for closed base LLMs as well as publicly available ones. 

Interestingly, this renormalized size tracks more closely anecdotally reported base LLM performance (excellent performance of T5 for its size, under-performance of MoE models despite over a trillion parameters), as well as anecdotally reported public perception of conversational agents (BARD/LaMDA and Galactica compared to ChatGPT and BingGPT), and notably the BIG-58 benchmark results for models they are available for \citep{PaLM2022GoogleJeffDean}.

While this rating is somewhat limited, given that for equivalent model size and dataset size, the performance is likely to be impacted by the quality and nature of the dataset, as well as the architecture of the model. However, we believe that it represents a more practical view of relative LLMs' capabilities. It would be interesting to see if the renormalized size indeed tracks the models' performance on task benchmarks.

% Overall, if we use the performance across a set of tasks compared to a human presented in \citet{PaLM2022GoogleJeffDean}, we see a strong correlation of performance with a combination of the model size and training dataset size, with 70B-parameter Chinchilla model performing very close to the PaLM model, despite being almost 8 times smaller, and only narrowly falling short of an average human across a benchmark of tasks presented in that paper. Once we apply the optimal 1:10 parameter-to-token scaling rule derived by \citet{ComputeOptimalLLMs2022Google} and limit the size of the models by 1/10th of the training datset tokens, we see that the "honest" model sizes still show an exponential-like growth in LOWESS regression, and that that "honest" size tracks tracks the models performance across 58 BIG-Bench task presented in \citet{PaLM2022GoogleJeffDean} (Fig.\ref{fig:honestsizes})

As such, the vast majority of LLM models still have room for further pretraining and likely would benefit from additional data injection, especially with data specific for potential downstream model applications, such as question answering or chain-of-thought explanation. A possible avenue of improvement is the creation of occasions by LLM developers for humans to write longer, more complex texts, for instance, literary stories competitions.

Overall however, as indicated by the pivot to conversational agents, the developers of LLMs seem to see the refinement in model architecture, fine-tuning, guided generation, and augmentation with auxiliary capabilities as the most likely sources of improvement in the short term rather than the model size alone.

\subsubsection{Better Multilingual Representation}

While the modern LLMs have emerging multilingual abilities \cite{UnpredictabilityOfLLMs2022Anthropic}, crawl-based training datasets used to train them are likely to be biased towards English as the predominant language of Internet resources. According to Alexa's rankings, among the top 10M websites, over 60\% are in English, with German and French only representing around 2.5\% of resources. While the linguistic bias likely present in global crawls is offset by the highly weighted Wikipedia dataset (2.5M articles in English against 0.8M and 0.7M in French and German, respectively), the other heavily weighted dataset for GPT family - websites pointed to by Reddit - is likely to be heavily biased towards English, given that the vast majority of the content on it being in English.

BLOOM model developers have specifically targeted resources in French, Spanish, and Portuguese to arrive at an almost complete balance between Romance languages and English in their training dataset \citep{Bloom2022HuggingFace}. Unfortunately, there is no indication that German and Germanic languages, except for English, were selected for a better representation in the BLOOM model.

Similarly, the GPT4 model developers, in their technical report \citep{GPT42023OpenAI}, showed a more limited performance degradation in other languages compared to English. It is, however, currently unclear now much impact such degradation would have on the model uses in the applications at scale. 

If the LLM-derived tools are to gain adoption across Europe, and in particular across Switzerland, more linguistically representative and varied models will be required. It is possible that the BLOOM family was not trained in a computationally-optimal fashion, in which case additional pretraining with more multilingual resources could be performed on it.

\subsubsection{Federated Learning, Privacy, and Byzantine Resilience}

A substantial issue for further scaling of LLMs is the depletion of available linguistical resources on the internet. GPT-3 initial dataset already contained the cleaned version of Common Crawl, leaving text sources such as emails, social media interactions, or internal documents. All of them are susceptible to contain private information. In addition to that, for the largest owners of such data (Google with Gmail and Google Docs, Facebook, Microsoft, ...), withheld as granting competitive advantage for their proprietary LLM training.

While diversification of more specialized LLMs is likely and is already happening in the conversational agents' space (Google search-oriented, Facebook interpersonal communication-oriented, ...), an alternative avenue is to perform on-premises federated learning with privacy-preserving aggregation rules. 

An additional advantage of deferentially private federated learning is its ability to eliminate personally identifiable information from the training dataset in a way that would be resilient to prompt engineering. This has strong implications for compliance with GDP and similar regulations. 

Research in the field of privacy-preserving ML published in 2022 suggests that while differential privacy decreases the efficiency of learning, it allows a better generalization \citep{DiffPrivacyInML2022Bogdan}. While the tradeoff between privacy and training efficiency is still not clear for LLMs, this direction has a strong potential to further advance LLMs' development.

Finally, in the case of Federated Learning adoption at scale, it could potentially include some malicious users that would attempt to poison LLMs, for instance, to influence their generation in their own interest. This problem is generally known as "Byzantine-resilience," and algorithms that resist it - as Byzantine-resilient \citep{ByzantineResilientML2017}. Research published in 2021 suggests that training large models that are simultaneously privacy-preserving and byzantine-resilient is problematic \citep{DiffPrivacyAndByzRes2021Rafael}, although research ongoing as of the date of this report suggests that there are ways to drastically increase the sizes of the models that can be trained in a Byzantine-resilient and privacy-preserving way. However, the degree to which it could be applicable to LLMs is yet to be determined. Similarly, it is unclear whether a major player in the domain would emerge and would be able to capture all the private information sources through network effects - as OpenAI/Microsoft joint operations could.

\subsection{Industry Business Model Trends and Impact on Governments}

% \andrei{The way I see this section is to provide an expert opinion about business needs, match it with existing offers and predict the market penetration in the short term. In turn, that would have implications for cyber-defense implications.}

% \andrei{Regulatory landscape as well? Or just mention it and say it's a rapidly evolving domain for which we do not have expertise to have an informed opinion on the topic?}

% \zach{The current section title 'Business Needs vs LLM Capabilities' is slightly different than what was outlined in Alain's mail 'Industry/business model trends and impact on governments' so please let us know if this fits the intended direction.}

% \andrei{Alain?}

\subsubsection{Impact on the Job Market}

We anticipate that the job market will be greatly impacted in the years to come by LLMs as many tasks and skills are now either within reach of direct automation or will massively benefit from adopting machine-assisted workflows. This could accelerate the ongoing trends of automation, leading to an era often referred to as hyper-automation. Though, rather surprisingly, the more creative-oriented jobs that were once thought to be safe from automation are the very ones at risk now. Indeed, it seems that advances in generative AI have surpassed those of autonomous robotic systems, which were once feared to be the greatest threat to large-scale workforce automation. Now it appears that aspects of the jobs of designers, journalists, and other creative writers will increasingly be automated using generative AI technology. 

Companies and public institutions will either be able to do much more with the same workforce or reduce the headcount for the same delivery. When it comes to corporate upskilling and training, employees can now learn to perform a wide range of tasks through natural language interactions with LLMs. Through assisted-programming tools such as GitHub’s Copilot, this even includes writing, debugging, or optimizing code in a variety of popular languages (e.g. SQL, Python, R). For this reason, the importance of technical skills such as programming may decrease for roles where domain knowledge is more important than technical capabilities (e.g. data analyst). This trend was already in motion with the rise of no-code and auto-ML tools but is now likely to drastically accelerate with the advent of advanced LLMs available at scale. 

However, it is unlikely that LLMs fully replace programmers, given the need for expertise to validate the outputs and develop code that is too niche to be well-represented in  the LLM training dataset. Instead, we are likely to see an additional step up the abstraction ladder, in the same way as with the transition from hardware switches to binary, then to Assembler, then to procedural programming languages, then to object-oriented and functional ones, and finally to domain-specific modeling languages.

\subsubsection{Enterprise Landscape}

The enterprise landscape of LLMs is likely to consist of three types of players, ranging from the upstream creators of proprietary LLMs (\textit{providers}), to the midstream developers of fine-tuned variants or custom applications (\textit{integrators}), and finally to the downstream companies and individuals using the services (\textit{users}). 

\begin{itemize}
    \item \textbf{Providers:} Training LLMs from scratch is a tremendously expensive undertaking that is likely to restrict many new entities from entering the market. As of now, only a handful of organizations exist which have successfully trained such large models. These include big technology firms (e.g. Google, Microsoft, Meta), AI research laboratories (e.g., AI21 Labs, Anthropic, Cohere, OpenAI), and open-source AI organizations (e.g. HuggingFace, EleutherAI). Private companies plan to commercialize their proprietary models either by using them internally to drive their core business needs or by licensing them as public-facing APIs (Application Programming Interface) for commercial and personal use. Open-source organizations, on the other hand, have made their models freely available to the public without any direct plans for commercialization. Just as with software, users will therefore have the choice between two types of models: licensed or open source.
    \item \textbf{Integrators:} Outside of the providers and the larger AI community, LLMs remain a niche and unexplored technology. Most existing companies have little to no internal expertise on the technical underpinnings or downstream applications of such models. We anticipate that experienced technology companies and AI-oriented consulting firms will seize this opportunity to function as integrators of this technology for other companies and end users. This may involve fine-tuning base LLM models for specific domains or tasks, the building of custom LLM-powered applications, or a combination thereof. As an example, Salesforce has published many advancements in the scope of using and adapting LLMs for particular use cases, which suggests they will be a competitive player in the integrator space.
    \item \textbf{Users:} Other companies are likely to adopt LLM-powered solutions for a variety of tasks, including search, customer service, and content creation. Individuals may also use these services to benefit from many of the same productivity gains and creativity aids as companies. 
\end{itemize}

\subsubsection{Enterprise Applications of LLMs}

For many sectors, leveraging the power of LLMs in the coming years will be a strong differentiator, either through extensive cost-cutting resulting in competitive pricing or through the augmented quality of their services.  

In our assessment here, we are not explicitly endorsing any of these applications of LLMs as being risk-free. Indeed, even if we limit ourselves to only security aspects of their adoption, there are many potential security vulnerabilities (see Section~\ref{sec:swiss_cyber_defense}), that have yet to be adequately addressed. Nonetheless, the competitive forces at play will likely push companies to develop and ship applications before solving the breadth of security concerns. The widespread adoption of LLM technology in the industry is likely to be driven by the following applications: 

\begin{itemize}
    \item \textbf{Search:} Paired with online search engines, LLMs enable a powerful new form of conversational search. This allows the user to ask questions in a more natural way, with the possibility of follow-up questions and personalized responses. The companies which adopt this technology early and successfully have an enormous potential to disrupt the future of search, which is perhaps the core technology powering the modern internet. For examples, see Perplexity AI (https://www.perplexity.ai/), You (https://you.com/), and of course, Microsoft’s revamped search engine Bing chat (GPT-4) - discussed in Section~\ref{subsec:binggpt} (https://www.bing.com/). 
    \item \textbf{Customer Service:} Companies can use LLMs to handle increasingly complex customer inquiries through chatbots or virtual assistants. In Switzerland, this can strongly empower the private banks, neo-banks, and wealth managers to handle client requests about their portfolios and performance, reducing costs and allowing the institutions to be more competitive internationally. Tailored LLM-powered agents could also enable insurance companies to answer customer questions on their premium and insurance contracts, providing immediate and personalized exchange.  
    \item \textbf{Content Creation:} LLMs and other generative AI technologies can assist companies in generating a variety of content, including social media posts, visuals, blogs, advertisements, and product descriptions. This can save companies time and resources while also ensuring a consistent brand voice across multiple business lines and geographies. For smaller companies and even individuals, this drastically lowers the upfront costs and barriers of content creation, thereby democratizing access. Microsoft's Office 365 Copilot is an example of such usage, focused specifically on documents within the Microsoft Office Suite \citep{MSOfficeCopilot2023}.
    \item \textbf{Personalization:} LLMs can help companies personalize their marketing efforts by generating product recommendations and tailoring messages to individual customers. 
    \item \textbf{Data Analysis:} Given their impressive text-to-code capabilities, LLMs can also serve as an in-between for users seeking to manipulate and analyze massive quantities of data. This can enable those with little to no programming experience to perform the typical roles of a data analyst. Instead of relying on pre-developed dashboards, companies can integrate LLM-powered solutions which more flexibly respond to questions, empowering data-driven decision-making within companies. Once again, Microsoft's Office 365 Copilot is an excellent example of such an application and the capabilities it could provide to users.
\end{itemize}

\subsubsection{Regulatory Landscape}

The regulatory landscape around LLMs – and, more broadly, generative AI systems – is a rapidly evolving domain. Some of the important topics of focus are the existence of copyright material in training data, the security concerns of leaking confidential data, and the liability of LLM providers for any harmful misuse. To proactively address these concerns, we strongly recommend the government gather informed expertise to guide the governance and decision-making process.

\subsection{Investment Trends}
While industry business model trends are indicative of general trends in the domain, the actual development of LLMs is critically dependent on the resources available to teams working on them. This means that investment trends are not to be neglected, given that major investments can radically affect what business trends are allowed to thrive and develop and which ones will fail to gain enough resources to succeed, despite a potentially promising future.

\begin{figure}[H]
\centering
\textbf{Funding amount and post-money valuations of generative models-related firms}\par\medskip
\includegraphics[width=0.95\textwidth]{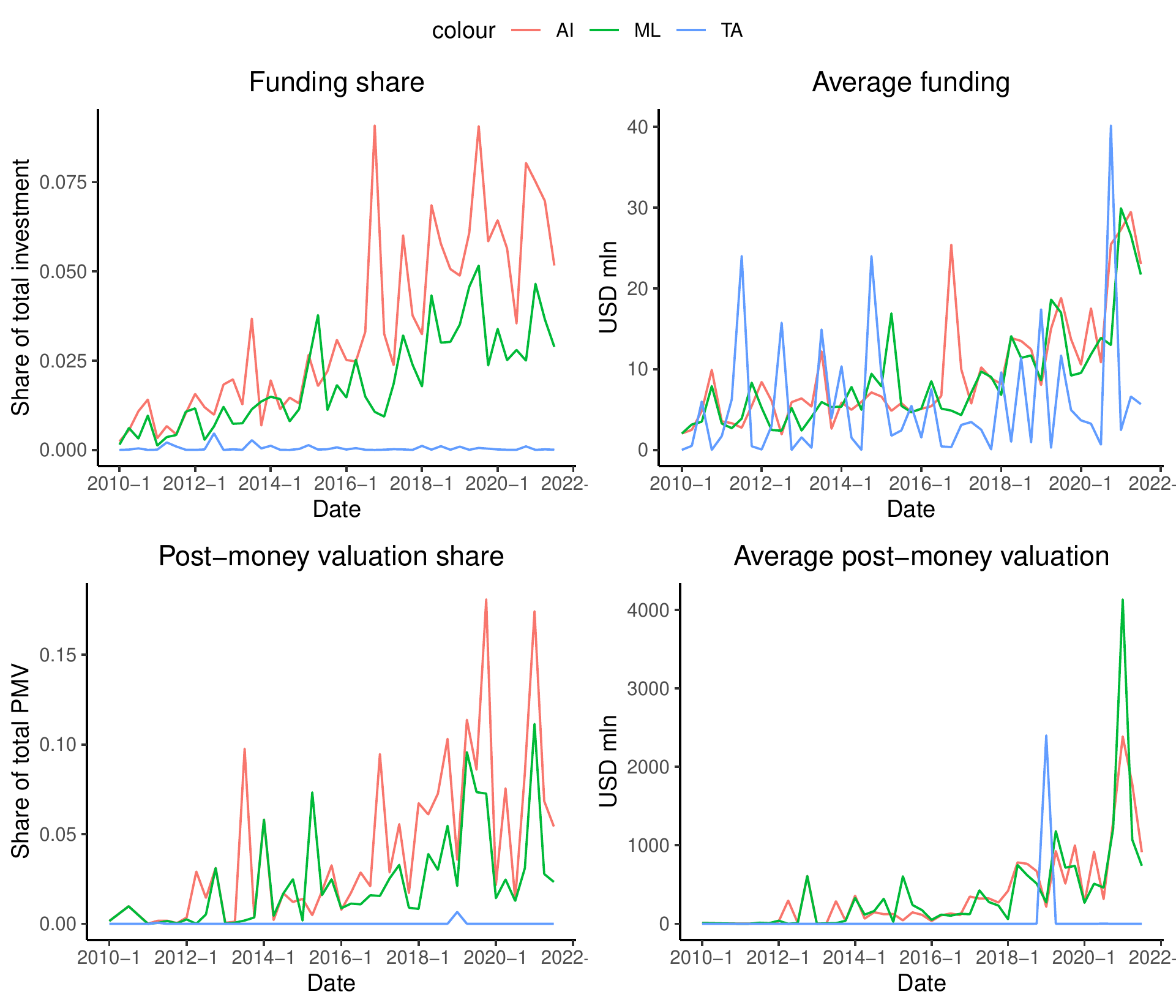}
\caption{\label{fig:ai_trends} Share over total funding amount and average funding amount and post-money valuations of private firms tagged with artificial intelligence (AI), machine learning (ML), and text analytics (TA). Data is from Crunchbase, the frequency is quarterly, and the period is Q1-2010--Q2-2022}
\end{figure}

Since LLMs' performances have become public, mainly through ChatGPT awareness, a large portion of the media releases cover the investment amount that LLMs developers have raised or will. One anecdotal but striking evidence is Microsoft's promise to renew and increase its stake in OpenAI, with a 10 bln USD investment series \citep{Bloomberg2023}.
To put these anecdotal values in context, we use Crunchbase \citep{Crunchbase2022}, a venture capital-related database, to retrieve information about funding round sizes and subsequent post-money valuations of firms active in the fields of artificial intelligence (AI), machine learning (ML), and text analytics (TA).\footnote{To minimize the effects of missing observations at the beginning of the sample, but also to avoid the effects of the 2008 global financial crisis, we restrict the analysis to the 2010-2022 period. In addition, this decade is consistent regarding the private equity investment boom and corresponds to the sector activity.} In the top Panels of Figure \ref{fig:ai_trends}, we depict the share of funding of each sector over the total funding recorded by Crunchbase, as well as the average funding from all series and debt financing. The AI and ML sectors' funding shares have increased from the order of magnitude of a basis point, to up to 7.5\% for the AI sector, over the last decade. The average funding increase was threefold over the period, with visual evidence for a significant increase from 2018 onwards. Conversely, given the restricted number of observations per quarter, it is challenging to uncover a trend in the average funding of the TA sector, which is also dwarfed by the overall ML and AI sectors.\footnote{The spike in average funding in Q4-2020 corresponds to a USD 100 mln series A targeting Blip, a Brazilian API/communication company\cite{WarburgPincus2020}.}
The share of post-money valuations in the AI and ML sectors has also increased from virtually inexistent to up to 15\% of the total valuations reported by Crunchbase. However, this sharp increase is also accompanied by significant volatility, with a sector valuation share likely to drop from 15\% to less than 5\% in one quarter. Although the number of observations also depends on the number of funding events, this should affect all sectors together. We interpret these drastic changes by uncertainties regarding the investors' expectations in the AI and ML sectors. By the same token, the average post-money valuation of the AI and ML sectors follows very close paths, with a substantial increase in the aftermath of the COVID crisis. At its peak, the average company of our sample in the AI (ML) sector is valued at 4 (2.5) USD bln.

\begin{figure}[h]
\centering
\textbf{Distribution of natural logarithm funding amount targeting generative models-related firms}\par\medskip
\includegraphics[width=0.95\textwidth]{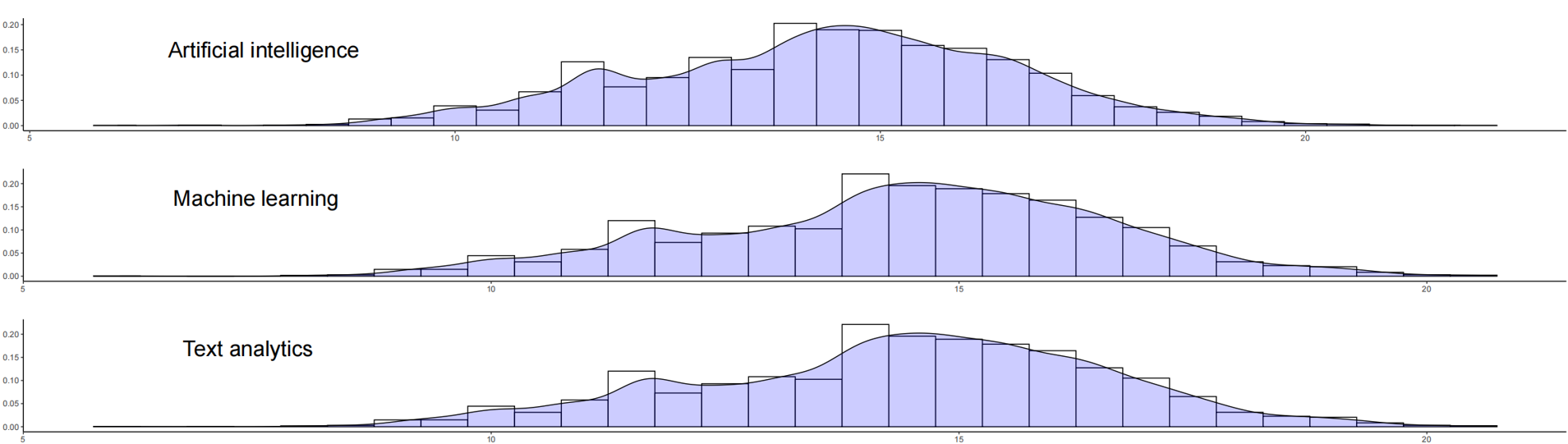}
\caption{\label{fig:ai_distr} Distribution of natural logarithm funding amount in private firms tagged with artificial intelligence (AI, top panel), machine learning (ML, middle panel), and text analytics (TA, bottom panel). Data is from Crunchbase, and the period is 2000--2022}
\end{figure}

% \newpage
The three panels of Figure \ref{fig:ai_distr} depict the log distribution of investment amount in each fields.  The distributions do not depart from the general private equity sectors, with significant outliers and some specific bins (generally around the one mln and one bln values) over-represented. For reference, whereas the previous maximum recorded investment in the AI space was three bln USD (natural logarithm value: 21.82 on the top panel), the not-yet recorded 10 bln USD investment of Microsoft in OpenAI would appear on a new right-bin of the distribution. 

% \newpage
Finally, as for the private equity market in general, investors and investees are for a vast majority present in the US, with 54\%, 65\%, and 56\% of the total funding amount targeting US firms for AI, ML, and TA, respectively. China comes second, with 24\%, 14\%, and 16\%, for the sectors, followed by the UK, Israel, Canada, and other developed countries making it to the top five. Switzerland gets half a percent of this share for AI and ML but does not reach the top twenty players for TA.

\begin{figure}[h]
\centering
\textbf{Distribution of funding amount targeting generative models-related firms per location}\par\medskip
\includegraphics[width=0.95\textwidth]{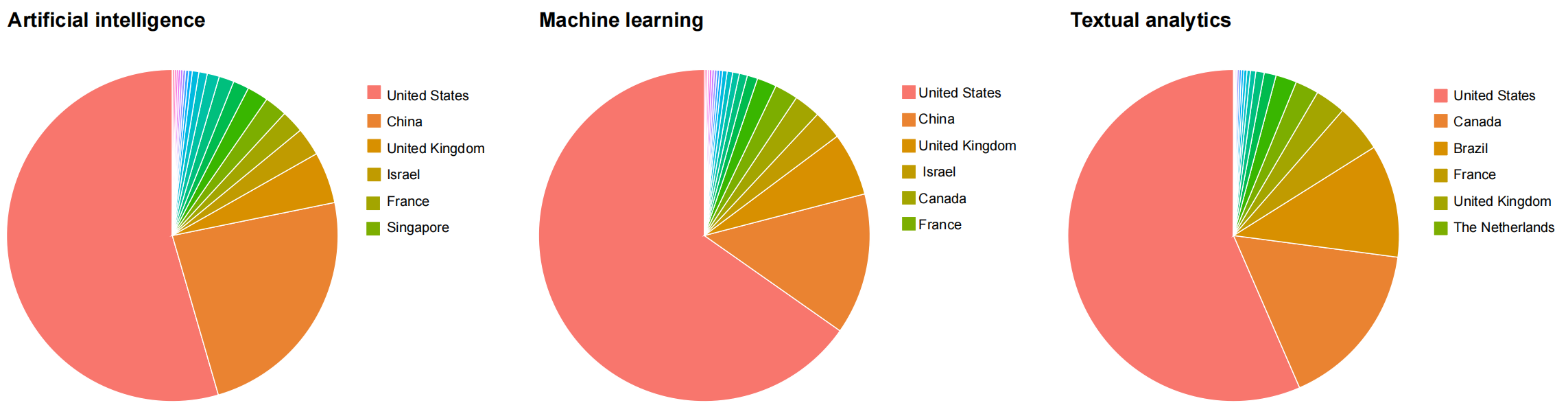}
\caption{\label{fig:ai_location} Share of funding amount in private firms tagged with artificial intelligence (AI), machine learning (ML), and text analytics (TA) by their 20 most important locations. Data is from Crunchbase, and the period is 2000--2022}
\end{figure}

\subsection{Public Attention Trends}

% The media have shown that there exists an increasing interest for the LLMs over the last months. It is important to measure effectively how the attention have evolve from the public view over time. For this, two proxies have been use to measure the public attention. The first one is Google trends. It allows to measure the interest of the population through search in the Google Engine. The second one is the OpenAlex database. It is an open source collection of scholarly documents. The Trends is measured by the number of citations to papers linked to a keyword. This proxy allows to visualize the attention's trend of the scientific community. Several keyword linked to the LLMs topic were measured in the proxies to get an idea of the evolution inside different communities.

While the expert opinion on business trends and investment analysis give a global insight into how the technology behind the LLMs and its adoption is likely to evolve, we need more immediate and direct proxies for adoption and specific technology development. Prior research indicates that the amount of attention to a component of a technical system is indicative of upcoming usage and speed of evolution of that technology, which we attempt to measure in this section \citep{daim_digital_2022, adner2002emergence, klepper1997industry, perez2010technological, rogers_diffusion_2010}. 

Specifically, we look at the general public attention towards technologies that allowed LLMs to emerge and rapidly progress through Google Trends -- a well-established, but relatively limited approach \citep{chumnumpan2019understanding, jun2018ten, woloszko2020tracking}. Aside from the methodological issue with Bing increasingly capturing the search volume of technically inclined users, bulk search volume is a trailing indicator of general interest in a specific technology. While it can be used as a proxy to forecast its adoption, it does not necessarily predict the speed of technological evolution, given that it is determined by small specialized communities. We attempt to forecast the speed of technology evolution by leveraging the OpenAlex database of scientific papers citations, annotated according to a topic ontology \citep{OpenAlex2023, priem2022openalex}.

Specifically, we focused on technologies that are generally cited by experts as key to current LLMs success, namely: \textit{Neural Language Model}, \textit{Deep Neural Language Model}, \textit{Attention}, \textit{Self-Attention}, \textit{Transformer Model}, \textit{Large Language Model}, \textit{Fine-Tuning}, \textit{Transfer Learning}, and \textit{Conversational Agent}.

\subsubsection{Google Trends}

\begin{figure}[h]
\centering
\textbf{Google Trends for specified terms}\par\medskip
\includegraphics[width=0.95\textwidth]{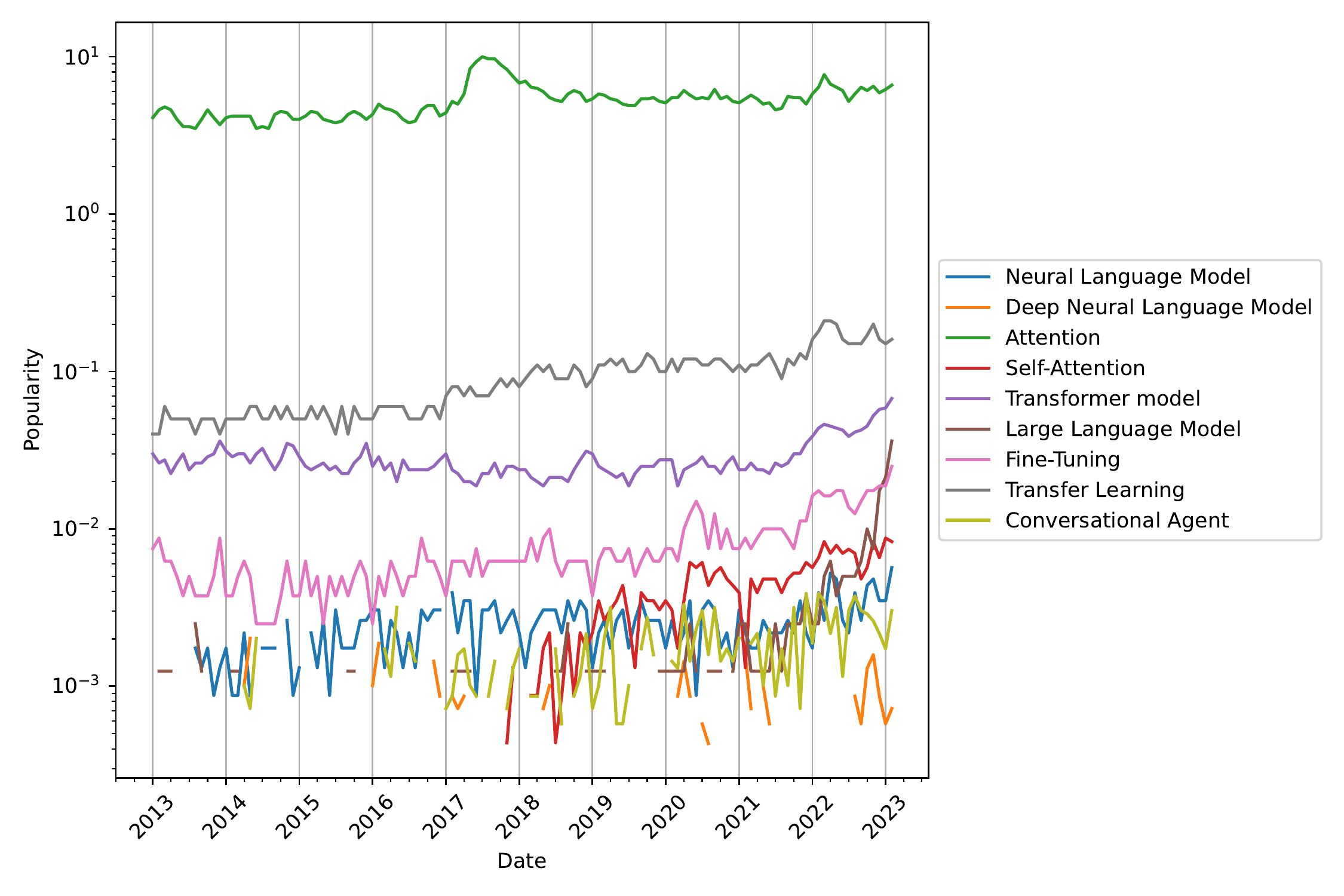}
\caption{\label{fig:proxy_google_trends} Attention capture by Google Trends, 2013--now}
\end{figure}

We used the \textit{g-tab}, an intermediate library to extract and normalize data from Google Trends \citep{gtab}. The resulting attention trends are presented in Figure \ref{fig:proxy_google_trends}. 

While all terms were retrieved, we observed substantial amounts of contamination in the less specific terms, notably "Attention", although "Transfer Learning" and "Transformer Model" were also likely affected until the public attention caught up to them in mid-2021. We observe a better trend accuracy for more specific terms, notably "Self-Attention," rising about a year after the release of the Transformer paper \citep{Transformer2017Google}, and "Large Language Models," starting from late 2021, as well as a more sustained interest in "Conversational Agents" from the same period. Finally, we observe a robust increase in "Fine-Tuning" starting from mid-2020, likely correlated to the publication of large Deep Learning models around that time period, both in Natural Language Processing and Computer Vision spaces. 

% Google Trends provides good results, but has some limitations such as the rounding of the popularity, lead to poor result for visualizing the attention. To solve these issues, the work done in the paper from \cite{gtab} was used. The Figure \ref{fig:proxy_google_trends} shows the evolution of popularity captured by Google. This provide a good overview, but since some keywords are generic terms and use in other topics as LLMs. It is important to take in account that some bias are included in the popularity of the keywords. An increases of attention by the public on LLMs since 2021 can be observe on all the keywords. Particularly on the general LLMs term where the popularity follows an exponential increase on the same period.

% via Google Trends Anchor Bank (g-tab), and the period is

\subsubsection{OpenAlex Citation Trends}

In the OpenAlex database of scientific articles, documents are linked to terms in a topic ontology that is internal to OpenAlex. Several technologies we were interested in did not have a corresponding ontology term and hence could not be investigated, indicating a first potential limitation of ontology-based databases for detailed technological forecasting.

\begin{figure}[h]
\centering
\textbf{Normalized OpenAlex citation trends for citation for specified terms}\par\medskip
\includegraphics[width=0.95\textwidth]{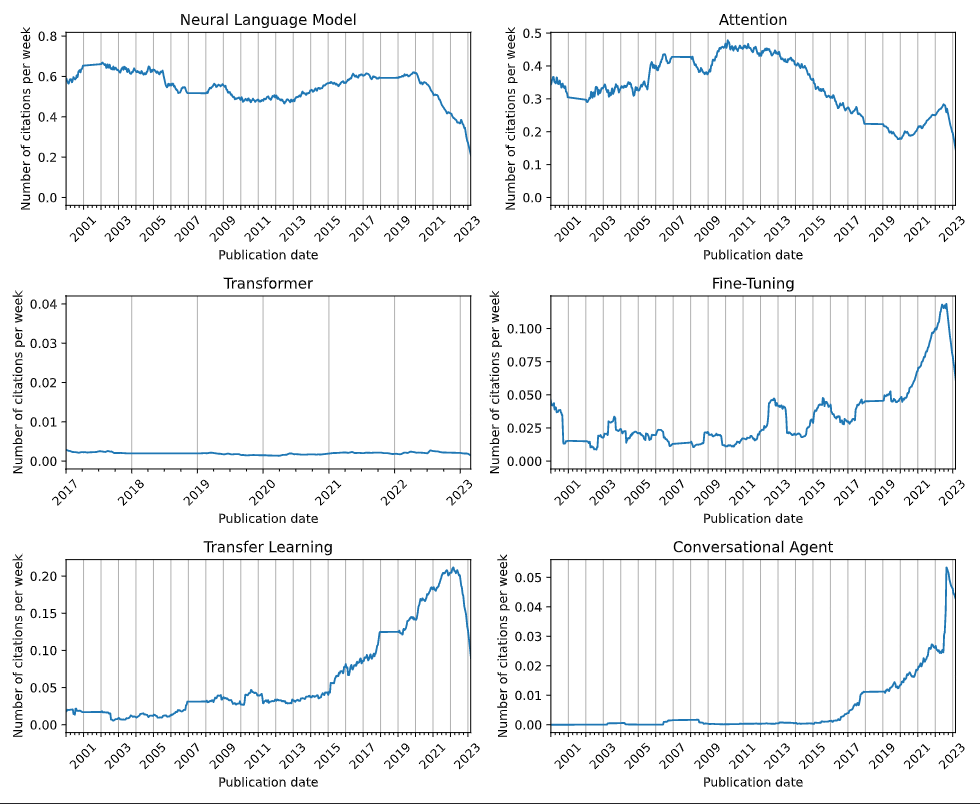}
\caption{\label{fig:proxy_openalex} Number of citations for papers, by the week of publication, filtered by internal ontology terms corresponding to specified terms. We limit The Transformer ontology entry to 2017-now to limit the contamination from unrelated publications.}
\end{figure}

An additional issue with the OpenAlex database was the delay in the data entry. Given that papers and citations are not added immediately, raw citation numbers tend to rapidly decrease within 3 years from the current time. To account for this bias, we normalized the number of citations per week to the total amount of citations per week across all the ontology terms in the OpenAlex database. While this did not solve the issue completely, it allowed for more interpretable trends. Finally, to minimize the contamination for the term "Transformer", we limited the date range for that term to 2017-now, given that 2017 is the year of publication of the seminal \citet{Transformer2017Google} paper. The final trends can be found in Figure \ref{fig:proxy_openalex}.

While "Fine-Tuning", "Transfer Learning" and "Conversational Agents" seem to agree with the expert opinion on the relevant papers and attention; "Attention" and "Neural Language Model" tend to anti-correlate with the timeline of the development of technology, whereas "Transformer" seems to be fully uncorrelated. The issue for that lack of correlation is likely contamination from other terms, whereas the anticorrelation likely corresponds to a mixture of contamination and semantic drift, with the "Neural Language Model" being gradually abandoned in the scientific literature in favor of more specific terms (eg "Transformer-Like" or "Full Self-Attention").

\subsubsection{Implications for Technological Forecasting}

The preliminary results from the two previous sections suggest that accurate technological forecasting for LLMs requires novel approaches. Specifically, those approaches would need to:
\begin{itemize}
    \item Not depend on a single ontology due to the delay in the emergence of a technology and a consistent name for it
    \item Account for delays in data recording introducing biases against more recent results
    \item Account for the semantic drift, eg realize that "Transformer-Like" and "Self-Attention" still designate "Neural Language Models"
    \item Accurately separate topics, such as Transformer Neural Language Model architecture and Transformers from unrelated fields
\end{itemize}

While we present the implications for LLM development forecasting specifically, the situation will likely be even more complex at the intersection of LLM technologies and technologies relevant to cyber-defense in general, suggesting a need for novel tools.

\section{Credit Attribution, Disclosures, and Acknowledgements}
\label{sec:metainfo}

\subsection{Corresponding Author}

\textbf{Corresponding author, lead contact, and technical contact}: Andrei Kucharavy

\subsection{Authors Contributions}

\textbf{Conceptualization:} A. Kucharavy, D. Percia David, A. Mermoud, V. Lenders

\textbf{Methodology, Software, Investigation, Data Curation, Visualisation, and Writing - Original Draft:}
\begin{itemize}
    \item A. Kucharavy - Introduction, GPT Family, Other Base LLMs, Fundamental Limitation of Generative Models, Implications for Swiss Cyber-Defense, Forecasting Short-Term Development- Expert Opinion.
    \item A. Kucharavy; L. Dolamic - Alternative Conversational Agents
    \item R. Sabonnadiere; Z. Schillaci: Forecasting Short-Term Development - Industry Business Model Trends and Impact on Governments
    \item L. Maréchal: Forecasting Short-Term Development - Investment Trends
    \item D. Percia David; M. Würsch: Forecasting Short-Term Development - Public Attention Trends
\end{itemize}

\textbf{Writing – Review \& Editing:} A. Kucharavy, L. Dolamic, Z. Schillaci, L. Maréchal, D. Percia David

\textbf{Supervision \& project administration:} V. Lenders, A. Mermoud, A. Kucharavy, D. Percia David

\textbf{Funding acquisition:} V. Lenders, A. Mermoud, A. Kucharavy, D. Percia David

\subsection{AI Tools Usage}
The only generative AI tools used in the writing of this report were grammar correction and rephrasing models, namely Grammarly.

\subsection{Ethics statement}

The goal of this report is to provide the necessary background on the LLMs and their implications for cyber-defense, notably within the operational context of Switzerland. We believe that at the level of granularity provided, the report would be more beneficial to encourage research into generative LLM safety, notably within the cyber-security and cyber-defense domains. We do not believe that any information given in the report can give an attacker an advantage.

Regarding the common safety and alignment criteria, we provide an extensive background to reader allow the reader to understand the inherent limitations of LLMs, notably with regard to their counter-factuality, private information leakage, harmful content generation, harms of representation, and biased decisions, and user overreliance. Additionally, we provide several avenues for monitoring, forecasting, and mitigating unsafe AI research acceleration, especially in the context of cyber-security and cyber-defense.

As such, we believe that this report is generally beneficial to the general public and to the AI research field, which motivated its public release.

\subsection{Conflicts of Interest}
R. Sabonnadiere and Z. Schillaci are CEO and Head of AI respectively at Effixis, a company specializing in AI consulting and training. Other authors declare no conflict of interest.

\subsection{Acknowledgements}
This document is the result of a research project funded by the Cyber-Defence (CYD) Campus, armasuisse S+T. AK is funded through the Swiss Confederation DDPS Research Contract No 8203005271. LM is funded through the Swiss Confederation DDPS Research Contract No 8203004738. DPD is funded through his academic institution (University of Applied Sciences of Western Switzerland, (HES-SO Valais-Wallis)).

\bibliography{bibfile}

\end{document}